\documentclass[sigconf]{acmart}
\usepackage{soul}
\usepackage{url}
\usepackage[utf8]{inputenc}
\usepackage{caption}
\usepackage{tikz}
\usepackage{amsmath,amsfonts, bm}
\usepackage{algorithm}
\usepackage{algorithmicx}
\usepackage[noend]{algpseudocode}
\usepackage{mdframed}
 
\usepackage{graphicx}
\usepackage{textcomp}
\usepackage{booktabs} 
\usepackage{siunitx}
\usepackage{xspace,url}

\usepackage{color}
\usepackage{colortbl}
\usepackage{multirow}
\usepackage{multicol}
\usepackage{makecell}
\usepackage{tabularx}
\usepackage{listings}
\usepackage{graphicx}
\usepackage{caption}
\usepackage{enumitem}
\usepackage{adjustbox}
\usepackage{subcaption}
\usepackage{listings}
\usepackage{xcolor}
\usepackage[colorinlistoftodos]{todonotes}

\acmSubmissionID{\#1004}

\definecolor{airforceblue}{rgb}{0.36, 0.54, 0.66}
\newcommand{\rev}[1]{{#1}}

\newcommand{\ours}{AgentProg\xspace}

\newcommand{\ie}{\textit{i}.\textit{e}.}
\newcommand{\eg}{\textit{e}.\textit{g}.}

\begin{document}

\acmYear{2026}\copyrightyear{2026}
\setcopyright{cc}
\setcctype[4.0]{by}
\acmConference[MobiSys '26]{The 24th Annual International Conference on Mobile Systems, Applications and Services}{June 21--25, 2026}{Cambridge, United Kingdom}
\acmBooktitle{The 24th Annual International Conference on Mobile Systems, Applications and Services (MobiSys '26), June 21--25, 2026, Cambridge, United Kingdom}
\acmDOI{10.1145/3745756.3809245}
\acmISBN{979-8-4007-2027-7/26/06}

\title{AgentProg: Empowering Long-Horizon GUI Agents with Program-Guided Context Management}

\author{Shizuo Tian}
\affiliation{%
	\institution{Tsinghua University}
	\city{Beijing}
	\country{China}
}

\author{Hao Wen}
\affiliation{%
	\institution{Tsinghua University}
	\city{Beijing}
	\country{China}
}

\author{Yuxuan Chen}
\affiliation{%
	\institution{Tsinghua University}
	\city{Beijing}
	\country{China}
}

\author{Jiacheng Liu}
\affiliation{%
	\institution{Peking University}
	\city{Beijing}
	\country{China}
}

\author{Shanhui Zhao}
\affiliation{%
	\institution{Tsinghua University}
	\city{Beijing}
	\country{China}
}

\author{Guohong Liu}
\affiliation{%
	\institution{Tsinghua University}
	\city{Beijing}
	\country{China}
}

\author{Ju Ren}
\affiliation{%
	\institution{Tsinghua University}
	\city{Beijing}
	\country{China}
}

\author{Yunxin Liu}
\affiliation{%
	\institution{Tsinghua University}
	\city{Beijing}
	\country{China}
}

\author{Yuanchun Li}
\thanks{Corresponding Author: Yuanchun Li (liyuanchun@air.tsinghua.edu.cn)}
\affiliation{%
	\institution{Tsinghua University}
	\city{Beijing}
	\country{China}
}
\renewcommand{\shortauthors}{Tian et al.}



\begin{abstract}

The rapid development of mobile GUI agents has stimulated growing research interest in long-horizon task automation.
However, building agents for these tasks faces a critical bottleneck: the reliance on ever-expanding interaction history incurs substantial context overhead. Existing context management and compression techniques often fail to preserve vital semantic information, leading to degraded task performance. We propose \ours, a program-guided approach for agent context management that reframes the interaction history as a program with variables and control flow. By organizing information according to the structure of program, this structure provides a principled mechanism to determine which information should be retained and which can be discarded. We further integrate a global belief state mechanism inspired by Belief MDP framework to handle partial observability and adapt to unexpected environmental changes. Experiments on AndroidWorld and our extended long-horizon task suite demonstrate that \ours has achieved state-of-the-art success rates on these benchmarks. More importantly, it maintains robust performance on long-horizon tasks while baseline methods experience catastrophic degradation. Our system is open-sourced at \url{https://github.com/MobileLLM/AgentProg}.

\end{abstract}

\maketitle

\section{Introduction}

Building intelligent assistants that can operate mobile devices by following natural language instructions has long been a central goal of mobile systems~\cite{ulink, kite, autodroid} and artificial intelligence researchers~\cite{metagui, seq2act, miniwob++}. Graphical User Interface (GUI) agents aim to simulate human user operations on graphical interfaces (\eg, clicking, scrolling, and inputting text) to interact with devices and complete complex tasks within existing software ecosystems. 
\rev{Mobile GUI environments present unique challenges compared to web or desktop settings: (1) \textbf{severe partial observability} due to the constrained screen size of mobile devices as only a small portion of UI elements is visible, and critical device states are usually hidden from the screen, making the observability problem more significant than on web or desktop platforms; (2) \textbf{noisy  accessibility trees} as many mobile applications provide incomplete, misleading or even no accessibility tree support, while a complete DOM tree is often readily accessible in web environments, making symbolic program approaches that depend on structured elements unreliable; (3) \textbf{fragmented UI ecosystems}, where the system share panels, permission dialogs, or even interaction patterns are different across different OS versions and device manufacturers, challenging the generalization ability of agents.} 
Recent advances in large language models (LLMs) and vision-language models (VLMs) have shown promising results in this domain. Building upon these foundation models, researchers have developed rich agent frameworks that enable agents to successfully complete relatively simple tasks through interactive decision-making processes~\cite{auto-ui,yang2023appagent,autodroidv2,mobilerl,mobile-agent-v3,agent-s3,mobileuse}.

However, despite their success on simple tasks, existing methods face significant challenges when scaling to long-horizon scenarios—tasks that require dozens or even hundreds of steps to complete. The ability to complete such long tasks is usually more desirable, as they are often cumbersome and difficult for human users. For instance, an example long task is to review all events on today's calendar (including sending messages, creating notes, etc.) and generate follow-up steps on all scheduled items, or to navigate between multiple shopping and note apps to compare and record product prices. 
These challenges manifest in two critical dimensions. First, model performance degrades significantly on extended tasks, as language models struggle to maintain coherence and task-relevant reasoning over protracted sequences~\citep{liu-etal-2024-lost,an2024does}. Second, the computational cost grows substantially as the interaction history increases, with each decision requiring to process an increasingly long context.

Existing approaches attempt to address the context management problem through various techniques, such as context window~\cite{uitars}, summarization~\cite{androidworld, mobile-agent-v3} or hierarchical planning~\cite{agent-s2, coloragent}. While these methods partially alleviate the computational burden, they face a fundamental limitation: they lack a principled framework for determining which information is essential for future steps and which can be safely discarded. This often results in the loss of crucial state information, leading to task failure~\cite{recap,thread,mem1}. Moreover, existing methods demonstrate a critical gap in generalization—an agent's ability to solve individual atomic tasks does not directly transfer to solving compositional or iterative long-horizon tasks, revealing deeper limitations in their reasoning capabilities~\cite{verigui}. The core challenge in long-horizon task execution lies in effective context management during task execution. An agent must maintain a comprehensive understanding of the task progress and dynamically assess the evolving environment to make informed decisions. This requires task-aware context management to preserve the essential state while discarding irrelevant details.

Based on a pilot test of mobile GUI agents on long-horizon tasks (detailed in Section~\ref{sec:motivation}), we observed three key difficulties that hinder more accurate and robust task execution.
\begin{enumerate}
    \item \textbf{Task Planning and Decomposition.} Long-horizon tasks often involve composition, repetition, and recursion of multiple subtasks. Existing methods typically decompose tasks into sequences or trees~\citep{cot,tot}, but these structures are overly simplistic for complex scenarios, necessitating frequent re-planning that introduces overhead and instability. For instance, tasks like ``read today's todo list and complete each item'' cannot be fully planned upfront since the items are unknown initially, leading to repeated re-planning that increases error risks.
    
    \item \textbf{Interaction History Management.} Due to the limited context length of LLMs, agents must determine what information to retain or forget in the task execution process. Current approaches lack task-aware context management, struggling to preserve critical intermediate results (\eg, restaurant details across a reservation task) while discarding irrelevant navigation history. 

    \item \textbf{Environment State Comprehension.} Agents must maintain an accurate understanding about environment states throughout execution. Software environments are usually dynamic and partially-observable with hidden UI elements, unexpected state changes, and transient failures, which are hard to handle for agents in the long task process. For instance, without proper environment understanding, agents may incorrectly believe tasks are complete when operations actually failed, causing workflow derailment that becomes critical in long-horizon scenarios.
\end{enumerate}

Our inspiration to address these challenges stems from traditional programs, which are capable of executing extremely long tasks with limited memory capacity. A program is naturally a representation of task decomposition. Its ``context management'' strategy is the program itself, which seamlessly handles information retaining/discarding in the system heap and stack through the mechanisms of functions, threads, global/local variables, etc. In principle, the task execution process of a GUI agent can also be represented as a program, and the agent context can be maintained in the same way as the CPU memory during program execution.

Therefore, we propose \ours, a program-guided context management approach for long-horizon GUI agent tasks. \ours reframes the agent's interaction history—traditionally represented as a flat, ever-expanding sequence — as a program with variables and control flow. This program structure, generated based on the task requirements, provides a principled and task-aware mechanism for context management. By organizing information according to the program's structure, \ours can systematically determine which information to retain and which to discard at each step, enabling task-specific context management that preserves crucial details.

A key question in our agent-program analogy is how to handle environment dynamicity. Letting the agent follow a program in the task process may lead to poor adaptivity, since the programs are generated before observing the actual environment at each step. We introduce a domain-specific language named Semantic Task Program (STP) to handle this problem. Unlike traditional programs that are based on precise symbolic instructions (\eg, \texttt{name = getElement('user Name').text}), STP uses fuzzy, natural-language-style instructions (\eg, \texttt{get user name as \{name\}}) to construct the program. The STP instructions capture necessary information for interaction context management, while they are interpreted adaptively at runtime based on the actual environment state. Furthermore, we incorporate a Global Belief State module into the STP execution process to address the inherent partial observability of GUI environments, a characteristic absent in traditional programming paradigms.

We evaluate \ours on AndroidWorld~\cite{androidworld}, a challenging and widely-used benchmark for mobile device-control agents, and introduce an extended task suite that we construct based on AndroidWorld to specifically test agent performance on long-horizon scenarios. Our experiments reveal that while existing methods perform well on the original AndroidWorld benchmark, they exhibit substantial performance degradation on our extended task suite, highlighting their fundamental limitations in handling long-horizon reasoning. In contrast, our program-guided approach demonstrates robust performance across both benchmarks. The context organization ability of \ours also helps it achieve the new state of the art on AndroidWorld.

Our main contributions are summarized as follows:
\begin{enumerate}
    \item We identify three key challenges for GUI agents on long-horizon tasks, and propose program-guided context management as a principled approach to address these challenges.
    
    \item We design and implement \ours, a mobile GUI agent based on program-guided context management. We introduce Semantic Task Program and global belief state to address the dynamicity and partial observability problems in program-driven task execution.
    
    \item \ours has been evaluated on AndroidWorld and AW-Extend and achieved the new state of the art.
    
    
\end{enumerate}
 
\section{Related Work and Motivation}
\label{section:background}

\subsection{GUI Agent}

Graphical User Interface (GUI) agents are autonomous systems designed to interact with software applications through their visual interfaces, mimicking human user behavior to accomplish complex tasks. A GUI agent typically takes as input a natural language task instruction (\eg, ``Send an email to John about the meeting'') and the current screen state, which includes visual information (screenshots) and optionally structural information (UI element hierarchies such as accessibility trees or HTML DOMs)~\citep{nguyen2025guiagentssurvey}. The agent outputs a sequence of actions such as clicking on specific UI elements, scrolling, typing text, or navigating between applications to accomplish the specified task within existing software ecosystems.

The decision-making process of a GUI agent can be formalized as a partially observable Markov decision process (POMDP)~\citep{1978pomdp}. At each timestep $t$, the agent observes the current screen state $o_t$ and selects an action $a_t$ based on the task goal $g$ and the interaction history $h_{<t} = \{(o_0, a_0), \allowbreak (o_1, a_1), \dots, (o_{t-1}, a_{t-1})\}$. The environment then transitions to a new state, providing observation $o_{t+1}$. This process continues until the task is completed or a maximum number of steps is reached. The core challenge lies in how agents manage the ever-growing interaction history $h_{<t}$ while maintaining accurate understanding of the current environment state.

\subsection{Related Work}
\textbf{GUI Agents with Foundation Models.}
Recent advances in large language models and multimodal large language models have significantly enhanced their capabilities in language understanding and cognitive processing, enabling effective interpretation of human instructions, detailed planning, and complex task execution~\cite{deviceLLMsurvey}. Early approaches combined LLMs with domain-specific knowledge to achieve smartphone automation~\cite{autodroid,yang2023appagent}, while subsequent methods introduced refined control mechanisms for mobile applications through visual perception and multimodal reasoning~\cite{mm-navigator,yang2023appagent,zhang2024ufo}. Vision-centric approaches have been proposed that identify and locate UI elements directly from visual inputs without relying on XML metadata or system accessibility trees~\cite{cheng2024seeclick,zheng2024seeact,gou2024uground,wu2024atlas,mobileagent,coloragent,mobileuse,wang2025uitars2}.

\textbf{Context Management for GUI Agents.} Managing interaction history is crucial for GUI agents to maintain task progress within token budget constraints. Existing approaches include \textbf{sliding window} methods that retain only the most recent $N$ observations~\cite{uitars}, \textbf{summarization} methods that condense each step into minimal tokens~\cite{androidworld,chain-of-memory,tan2024cradle}, and \textbf{hierarchical planning} methods that decompose tasks into sub-task sequences~\cite{mobileuse,mobile-agent-v3, agent-s3,zhang2024ufo,mozannar2025magentic}. However, these methods face limitations: sliding window incurs increasing context, summarization lacks explicit planning for complex tasks and may lose critical
early information, and hierarchical planning introduces overhead requiring multiple re-planning rounds and also faces with the challenge of forgetting important information.

\textbf{Program-Based GUI Agents.} 
While standard GUI agents utilize a fixed set of programmatic APIs (\eg, \texttt{click}, \texttt{type}, \texttt{scroll}) as their action space, they typically treat these low-level APIs merely as atomic tool invocations. Recently, a distinct paradigm shift in LLM agents has emerged from using rigid, discrete tool calls to employing executable code as the unified action space for agents~\cite{codeact, autoiot, li2025websailor, code_as_policy}. To bridge the gap between low-level actions and high-level code generation, recent GUI agents have empowered agents to proactively construct high-level functions, encapsulating primitive operations into meaningful skills~\cite{wang2025inducingprogrammaticskillsagentic, os_copilot}. Other approaches establish a persistent program context or terminal environment, enabling the agent to engage in continuous interaction through script synthesizing~\cite{autodroidv2, song2025coact, program-synthesis}. 
However, current task script generation methods~\cite{autodroidv2} typically require precise API calls and strict syntax adherence, relying heavily on extensive exploration of mobile apps to build application documentation. Consequently, these scripts are prone to failure when encountering unseen applications or when dealing with out-of-distribution scenarios where documentation is unavailable.

\subsection{Motivation and Challenges}
\label{sec:motivation}

\begin{figure}[htbp]
    \centering
    \includegraphics[width=\linewidth]{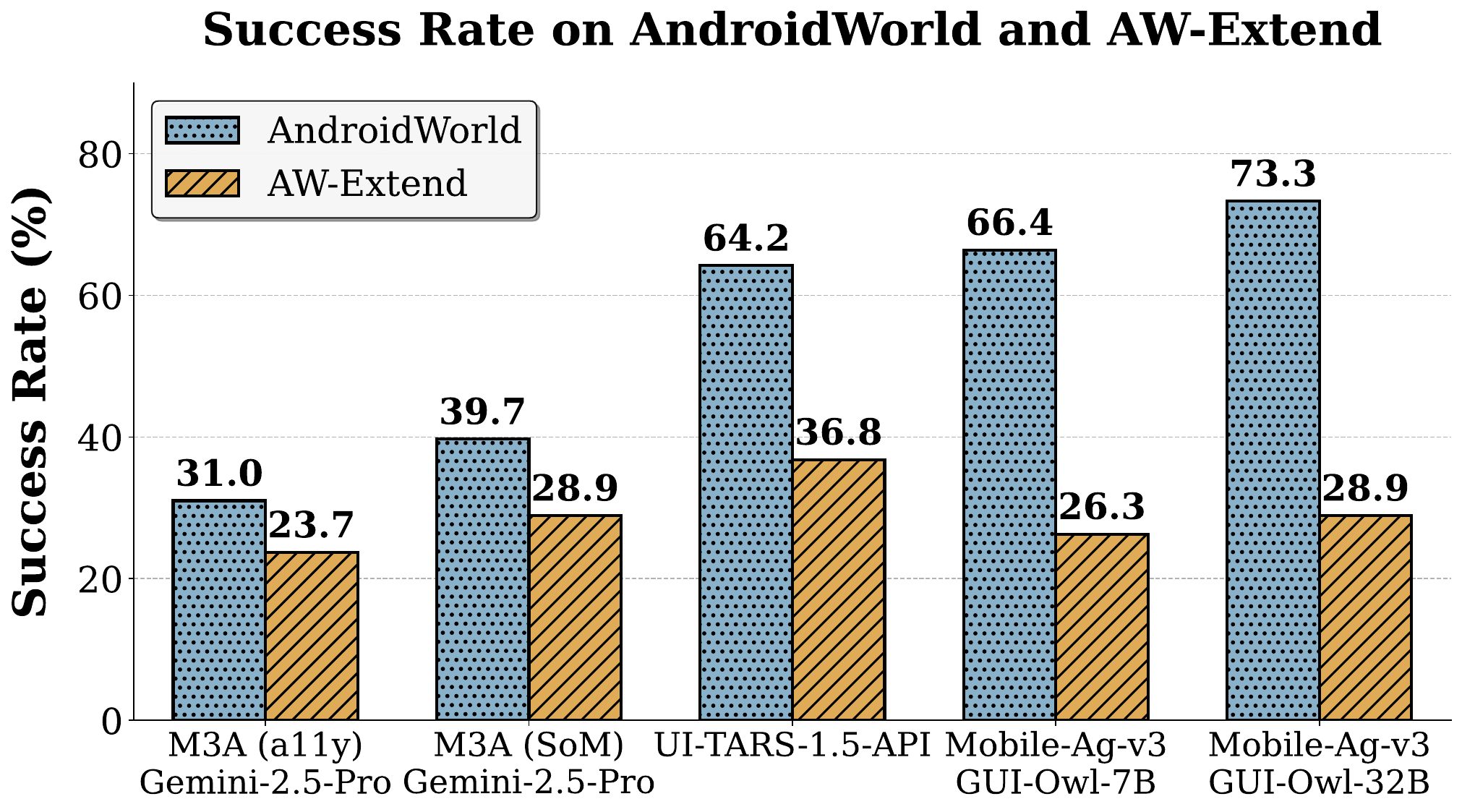}
    
    \caption{
    \textbf{Performance Comparison on AndroidWorld vs. AW-Extend.} 
    \textnormal{\textit{a11y} refers to the Accessibility Tree observation space; \textit{SoM} denotes Set-of-Mark; \textit{Mobile-Ag-v3} denotes Mobile-Agent-v3.}
    }
    \label{fig:extended_results}
\end{figure}

\begin{figure}[h!]
  \centering
  \includegraphics[width=\columnwidth]{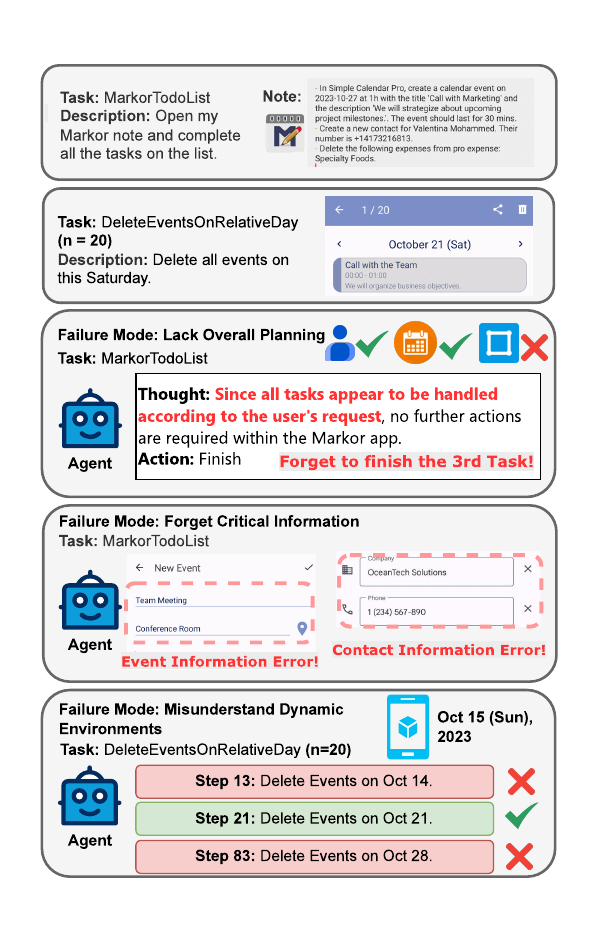}
    \caption{Failure mode in existing methods (Mobile-Agent-v3~\cite{mobile-agent-v3}, UI-TARS~\citep{uitars}, M3A~\citep{androidworld}) on AW-Extend.}
    \label{fig:pre_case_study}
\end{figure}

Despite the progress in GUI agents, existing methods face fundamental limitations when handling long-horizon tasks, which require dozens or hundreds of sequential steps to complete. To better understand these challenges, we construct AW-Extend, an extended task suite based on AndroidWorld~\cite{androidworld} that specifically tests agent performance on compositional and iterative tasks. Our extended suite comprises 19 tasks that are possibly common in realistic scenarios across two categories: \textbf{Compositional Tasks} that combine sub-tasks into a compositional task requiring critical information maintenance across applications, and \textbf{Iterative Tasks} that scale sub-task counts to $n=10$ or $n=20$ to test robustness and memory filtering capabilities. The detail of this task suite is illustrated in Section~\ref{sec:aw_extend}.

Our preliminary analysis reveals critical failure reasons in existing methods, as illustrated in Figure~\ref{fig:extended_results}:

\textbf{Lack of Overall Planning.} In the ``MarkorTodoList'' task, Mobile-Agent-v3 prematurely terminates after completing only the first two sub-tasks, mistakenly believing all three items from the note have been addressed, as shown in Figure~\ref{fig:pre_case_study}. This indicates a fundamental deficiency in holistic task planning and progress tracking.

\textbf{Failure to Remember Critical Information.} Despite employing context management schemes, existing agents fail to preserve task-relevant information at critical moments. In our tests, agents consistently pass incorrect parameters during sub-task execution because they do not retain all task items when initially opening the note, as shown in Figure~\ref{fig:pre_case_study}. Their context management mechanisms lack sensitivity to task-critical information.

\textbf{Poor Understanding of Dynamic Environments.} Our iterative tasks involving 10-20 sub-tasks expose fragility in handling unexpected situations—accidental app closures, dialog popups, or wrong operations. Moreover, due to the partial observability of GUI environments, changes in phone data are sometimes implicit, such as system settings, date and time, and in-app data. Existing methods struggle with such environmental variations, resulting in poor robustness during extended task execution. In our tests, the agent was tasked with deleting all events on this Saturday, but failed to correctly understand the date in the phone environment. Consequently, it deleted all Saturday events across different weeks, leading to task failure.

\section{Our Approach: \ours}
\label{sec:method}

\begin{figure*}[htbp]
  \centering
  \includegraphics[width=0.9\textwidth]{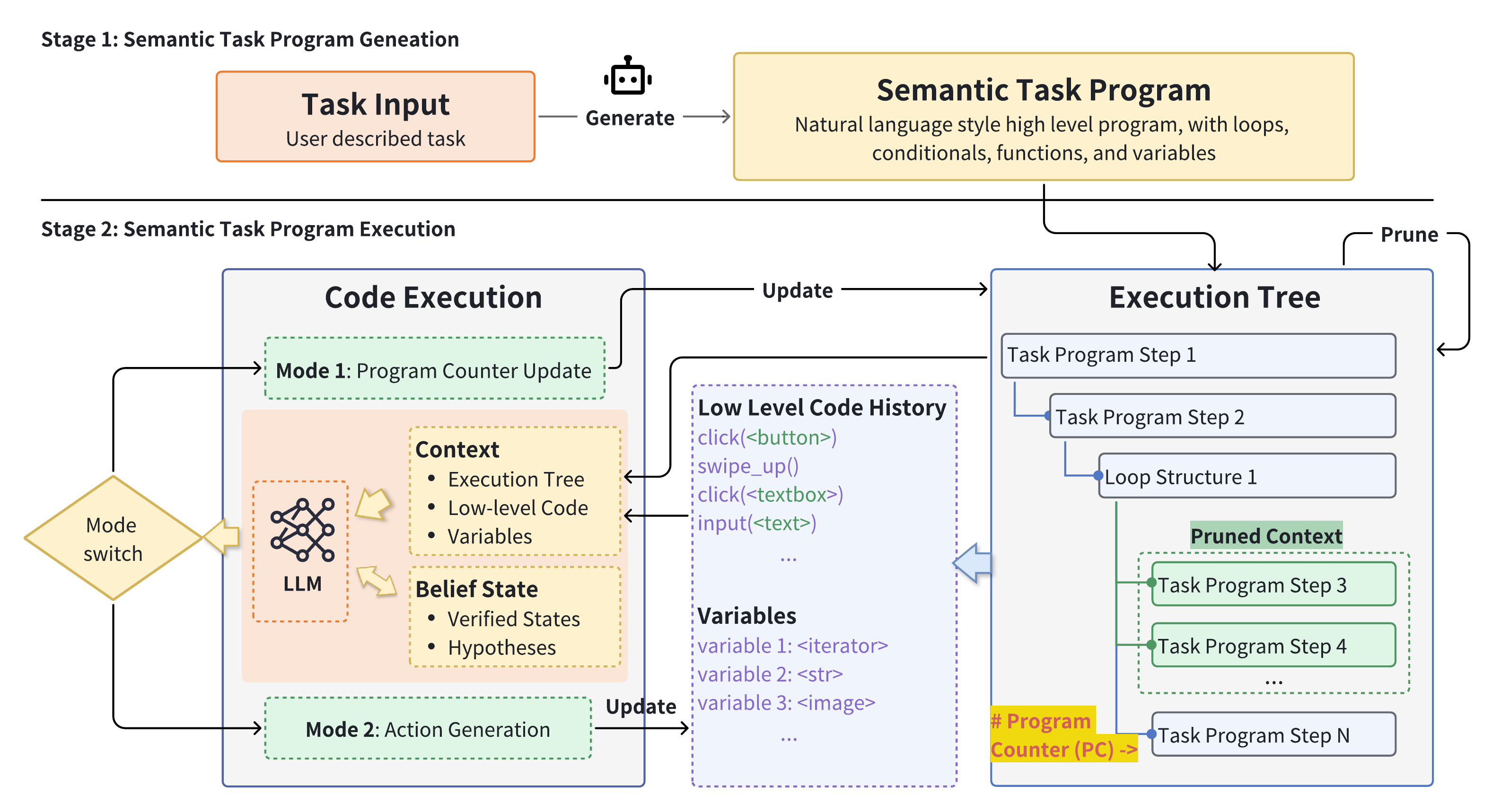}
    \caption{The workflow of \ours.}
    \label{fig:method}
\end{figure*}

We present \ours, a program-guided framework for long-horizon task execution, as shown in Figure~\ref{fig:method}. It addresses three fundamental challenges in agent-environment interaction. 
First, to tackle the challenge of planning and decomposing complex long-horizon tasks, we introduce Semantic Task Program (STP), a domain-specific language that extends natural language instructions with structured control flow (loops, branches, functions) and explicit variable operations. 
Second, to provide agents with appropriate context without redundancy, we develop a program-guided context management approach that operates along control flow-based pruning and data flow management. 
Third, to maintain accurate environment understanding despite partial observability and outdated beliefs, we propose a Global Belief State mechanism that continuously validates and updates the agent's hypotheses about environment mechanisms and hidden variables throughout execution.

Together, these components form a two-stage framework: Semantic Task Program Generation, where the agent performs global planning to create a structured program with identified critical variables, followed by Semantic Task Program Execution, where the agent incrementally interprets the program while maintaining belief state consistency and managing context efficiently. This design enables stable, efficient execution of long-horizon tasks without the context overflow and information loss problems that plague conventional approaches.

\subsection{Semantic Task Program: A DSL for Long-Horizon Task Planning}

We design Semantic Task Program (STP), a domain-specific language (DSL) that represents long-horizon tasks. Unlike rigid programming languages, STP balances expressiveness with semantic tolerance. It enables agents to plan and execute complex tasks effectively. More details and examples of STP are listed in Section~\ref{sec:dsl}.

\textbf{Design Principles.} STP is built on three core principles: (1) \textit{Natural Language Foundation}: Each program step is primarily expressed in natural language instructions, ensuring readability and maintaining generality across different environments; (2) \textit{Structured Control Flow}: The language incorporates essential programming constructs including loops, conditionals, and functions to express iterative and recursive task patterns concisely; (3) \textit{Explicit Data Flow}: Variables are used to explicitly pass information between steps, allowing the agent to identify and preserve critical information throughout task execution.

\textbf{Syntax and Constructs.} An STP consists of a sequence of steps, where each step can be: (1) \textit{Action Step}: A natural language instruction describing what should be accomplished, such as ``Open the note `todo\_list.md' in the Markor application''; (2) \textit{Variable Declaration and Assignment}: Explicit recording of critical information using variables enclosed in braces, \eg, ``record them as \{task\_list\}''; (3) \textit{Control Flow Statements}: Loop constructs, conditional branches, and function definitions that organize execution flow, \eg, ``Iterate over \{task\_list\}''; (4) \textit{Variable References}: Using declared variables in subsequent steps, \eg, ``for each \{task\_item\}, ...''.

\textbf{Semantic Tolerance.} A key distinguishing feature of STP is its semantic tolerance. Current task script generation methods~\citep{autodroidv2} require precise API calls and strict syntax. Consequently, they are highly sensitive to environment dynamics and software updates. Semantic Task Program allows for variations in execution details. The natural language instructions provide high-level guidance without being tightly coupled to specific environment implementations. This tolerance enables the agent to adapt to different environments and handle unexpected situations during execution, while the structured control flow ensures the overall task logic remains intact.

\rev{\textbf{Comparison to Symbolic DSLs.} Symbolic DSLs (\eg, AutoDroid-V2) often suffer from syntax errors and require extensive app-specific knowledge~\cite{autodroidv2}. When mobile applications provide poor or even no accessibility tree support, those methods are prone to failure. This critical limitation on mobile platform motivates our natural-language-style design, where STP leverages LLM's strong semantic priors, ensuring robust generalization via visual observations without rigid syntax constraints or limitations of accessibility tree.}

\subsection{STP Generation and Execution}

\ours aims to enhance agent performance in long-horizon tasks through a two-stage framework for memory management as Figure~\ref{fig:method} shows: 1) \textbf{Semantic Task Program Generation}: The agent generates a globally planned Semantic Task Program based on task requirements, where the program extends natural language with support for control flow constructs (\eg, branching, loops) and variable operations (declaration, assignment); 2) \textbf{Semantic Task Program Execution}: The agent incrementally interprets and executes the Semantic Task Program, taking environment state and current program state as input, and producing actions while updating the program state.

\textbf{Generation Phase.} The agent plans a feasible program according to specifications. It uses control flow to orchestrate execution and identifies critical variables for preservation. The natural language components ensure the program remains general and robust. 

\textbf{Execution Phase.} Instead of rigidly executing pre-defined scripts, \ours adopts a \textit{dynamic instruction grounding} paradigm. The Semantic Task Program serves as a high-level guide, where \ours acts as a runtime interpreter that translates the current high-level program instruction into executable low-level actions based on the real-time environment state. This mechanism decouples the task logic from specific execution details, preventing context overflow while maintaining flexibility.

Specifically, \ours operates in two alternating modes: 

\textbf{1) Action Generation.} 
In this mode, the agent focuses on \textit{instantiating} the high-level intent of the current program step into concrete execution logic.
Based on the instruction pointed to by the Program Counter (PC) and the current GUI observation, the agent generates a snippet of low-level Python code to interact with the environment. 
\ours utilizes LLM to ground abstract natural language instructions into a sequence of atomic Python APIs (\eg, \texttt{start\_app(<app>)}, \texttt{click(<element>)}, \texttt{swipe\_down()},
\texttt{input}\\\texttt{(<text>)}).
Crucially, this allows the agent to dynamically adapt the low-level operations (\eg, handling pop-ups or layout changes) while adhering to the high-level plan. The agent records these executed Python scripts to track data flow, while repetitive execution histories (\eg, inside loops) are folded to save context.

\textbf{2) Program Counter Update.} After the generated Python code is executed and the environment updates, the agent switches to this mode to manage the control flow. 
It decides how to move the Program Counter—whether to advance to the next step, iterate a loop, or jump to a branch—based on the execution result of the previous step. \rev{The Program Counter is dynamically determined by the LLM rather than following a deterministic transition, enabling the agent to skip, repeat or branch based on runtime observations.}

These two modes alternate strictly: \ours translates the current instruction into Python code (Action Generation), executes it, and then decides where to go next in the program (PC Update). Throughout this process, \ours maintains a structured context containing the \textit{static} program plan and the \textit{dynamic} variables and low-level history, ensuring all decisions are globally consistent yet locally adaptive.

\subsection{Program-Guided Context Management} 

During the Semantic Task Program Execution phase, \ours adopts a program-guided approach to manage the context input for the agent. This mechanism addresses the trade-off between information retention and context length. By leveraging the explicit structure of the STP, \ours acts as a filter: it retains execution-critical information while systematically discarding redundant history. This management operates along two dimensions: \textit{control flow pruning} and \textit{data flow persistence}.

\begin{figure*}[!htb]
    \centering
    \includegraphics[width=\textwidth]{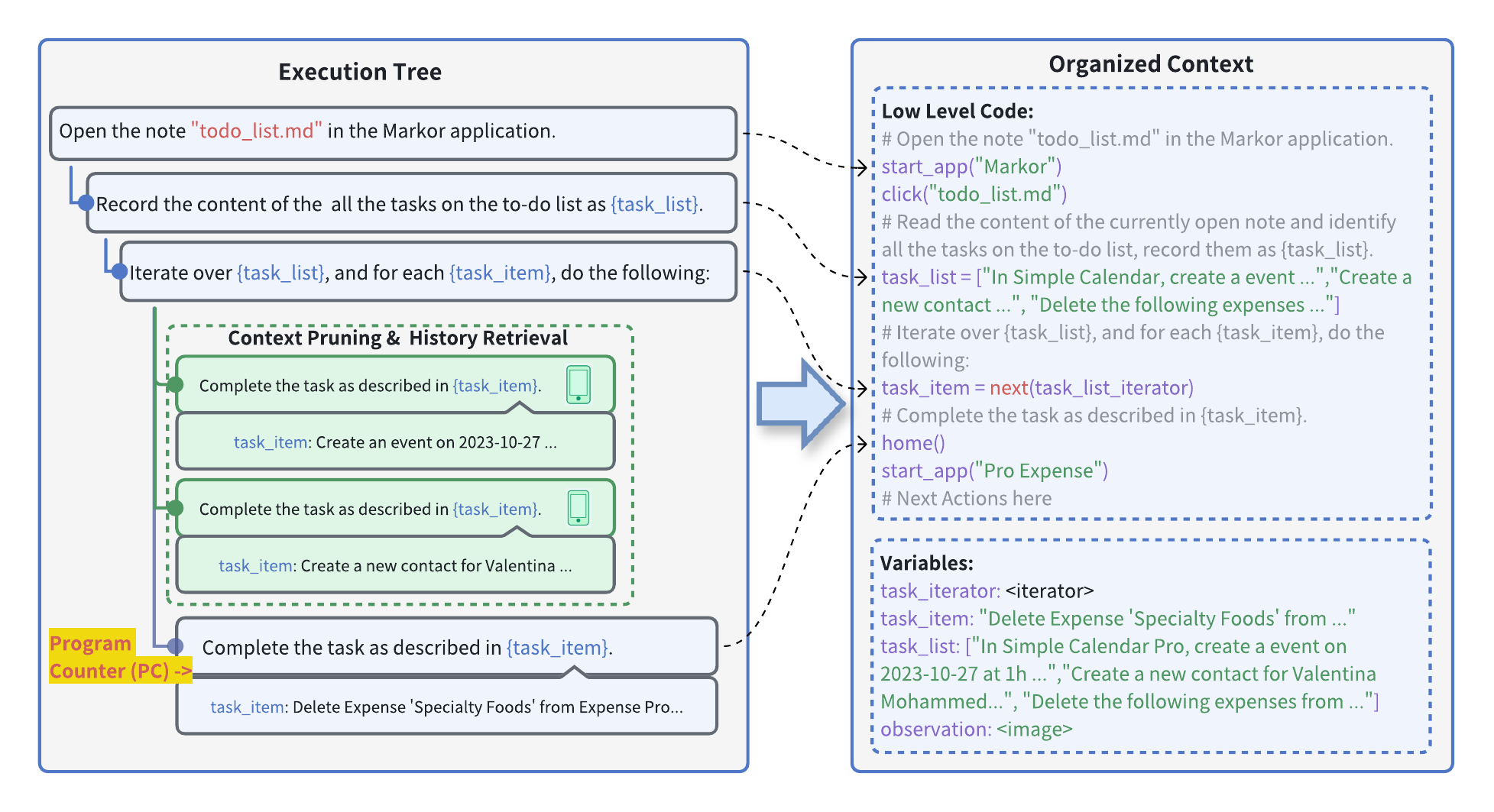}
    \caption{Program-guided context management with context pruning, history retrieval and variable management.}
    \label{fig:execution_tree}
\end{figure*}

\subsubsection{Execution Tree-Guided Context Pruning.} 
To manage the control flow context, we organize the STP execution history into a dynamic tree structure. As illustrated in Figure~\ref{fig:execution_tree}, each executed step expands a corresponding node in this tree. 

\textbf{Tree Construction Rules.}
The node expansion logic is strictly bound to the program structure. We categorize nodes into three types:
\begin{itemize}
    \item \textit{Sequential Nodes:} If a step follows a linear sequence, it becomes a direct child of the previously executed node.
    \item \textit{Conditional Nodes:} For branching statements (\eg, if-else), only the node corresponding to the \textit{actually executed} branch becomes a child of the condition node. Unexecuted branches are never added.
    \item \textit{Loop Nodes:} For loop structures, each iteration expands a new loop statement node. These are added as children of the loop entry point, preserving the iteration history hierarchically.
\end{itemize}

\textbf{Path-Based Pruning.}
During inference, the agent does not see the entire raw history. Instead, we compute the \textit{active path} from the root node to the currently executing node. Only the step information and observations along this active path are retained in the context; all other branches and completed loop iterations that are no longer relevant to the current state are discarded. 

This approach significantly reduces context overhead, especially in tasks involving complex logic. It prevents the agent from being confused by redundant information (\eg, stale states from a previous loop iteration or irrelevant actions from a non-taken branch), allowing it to focus solely on the current logical path.

\subsubsection{Program-based Historical Step Retrieval}
While pruning removes irrelevant context, \ours also employs a retrieval mechanism to enhance stability for iterative tasks. 
When the execution revisits a specific program step—a common scenario in loops or recursive segments—\ours activates a retrieval mechanism. Unlike generic similarity search, this mechanism uses the unique program step ID to precisely identify preceding executions of the \textit{exact same step}. 
The operational outcomes of these past iterations—including actions taken and resulting state changes—are selectively injected into the current context as references. This allows the agent to ``remember'' its own recent success or failure patterns. By providing this structured historical awareness, \ours significantly improves consistency in long-horizon repetitive tasks, mitigating the risks of oscillation or repeated errors that often plague agents lacking step-specific memory.

\subsubsection{Data Flow and Variable Management}

Beyond control flow, \ours ensures the persistence of critical information through explicit variable management. This addresses the challenge where purely context-based agents often lose track of key data (\eg, a phone number retrieved 50 steps ago).  
During the Semantic Task Program Generation phase, the agent explicitly identifies and declares essential variables. These variables—such as extracted todo items, reservation times, or user preferences—act as ``anchors''. They are mandated to persist throughout the execution, ensuring that task-critical data is never inadvertently discarded by the pruning mechanism. 

During the Execution phase, the agent retains the flexibility to define auxiliary variables via Python code. This programmatic manipulation compensates for the limitations of raw LLM reasoning. For instance, an agent can extract multiple items into a list variable, iterate through them systematically, and use counters to track progress. 
This combination of \textit{declarative planning} and \textit{imperative execution} ensures robust information flow: critical data remains accessible, while the internal state adapts dynamically to evolving task requirements.

\subsubsection{Advantages of Program-Guided Context Management. } 
\ours binds context directly to the program's execution structure, effectively addressing these limitations. 
From a \textbf{control flow} perspective, the STP captures repetition and branching patterns upfront. This allows the agent to systematically prune irrelevant branches and loop iterations without the need for error-prone re-planning cycles.
From a \textbf{data flow} perspective, explicit variable declaration serves as an information anchor, ensuring that critical data (\eg, user preferences or extracted items) persists throughout execution regardless of context length. 
This dual mechanism enables \ours to maintain a compact, noise-free context while guaranteeing the persistence of task-critical information.

\subsection{Global Belief State} 
\begin{figure}[htbp]
  \centering
  \includegraphics[width=\columnwidth]{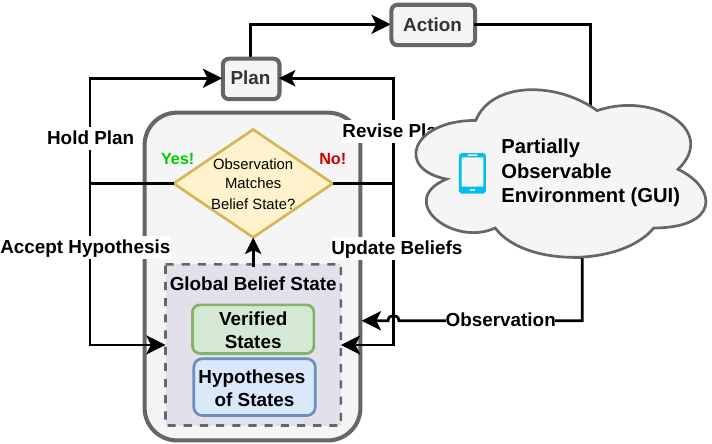}
    \caption{Dynamic global belief state management in partially observable environments.} 
    \label{fig:global_belief_state}
\end{figure}

While the \textit{Action Generation} phase produces executable Python scripts, ensuring their faithful execution requires active monitoring. The \textbf{Global Belief State} integrates directly into this low-level execution loop, acting as a \textit{runtime verifier} that checks if the assumptions made during code generation remain valid as each atomic command executes.
This validation is critical because mobile GUI environments are inherently partially observable. Critical system states—such as {clipboard content}, {navigation back-stack}, or {off-screen selected items}, etc.—are often invisible in the current screenshot but decisive for future actions. 

Inspired by the Belief MDP framework~\citep{1998beliefstate}, we introduce a \textbf{Global Belief State} mechanism to bridge this gap. 
It operates as a runtime monitor through a continuous \textit{Predict-Verify-Align} cycle:
\textbf{1) Maintain Hypothesis (Predict):} 
Unlike passive memory, the agent maintains an active set of hypotheses about the current environment status, including the UI context and data validity.
\textbf{2) Runtime Verification (Verify):} 
During the execution of low-level Python codes, \ours continuously compares the real-time visual observation against its belief state. 
\textbf{3) Anomaly Correction (Align):} 
If the observation contradicts the belief, \ours identifies a \textbf{Belief-Reality Gap}. It immediately marks the previous state as invalid (\eg, ``Context Lost''), and switches from execution mode to a recovery routine (\eg, navigating back to the target app) to realign its belief with reality.



\textbf{Case Study.}
Consider a scenario where an agent is filling out a contact form. 
\textit{Initial Belief:} ``Editing contact `John Doe'; Name and Email fields are filled.'' 
\textit{Event:} Due to an accidental touch or crash, the app exits to the home screen.
\textit{Without Belief State:} A traditional agent might hallucinate that the step succeeded and attempt to click a "Save" button that no longer exists, leading to a cascade of errors.
\textit{With Belief State:} \ours detects the conflict: the visual observation (Home Screen) refutes the belief hypothesis (Editing View). It immediately invalidates the "Name/Email filled" status, updates the belief to "Form Context Lost," and generates a corrective plan to restart the app and re-fill the information. This mechanism ensures that the agent never proceeds on false premises.
\section{AW-Extend Benchmark}
\label{sec:aw_extend}
To systematically evaluate agent performance on the long-horizon challenges identified above, we constructed \textbf{AW-Extend}, an extended task suite that builds upon the original AndroidWorld benchmark. By adapting the existing AndroidWorld evaluation and verification system, we ensure our extended tasks maintain the same rigorous automated assessment standards while specifically targeting long-horizon scenarios. 


\rev{Existing mobile GUI benchmarks~\citep{androidworld,androidlab,mobile-agent-bench} primarily focus on evaluating isolated functionalities within individual applications, lacking coverage of realistic complex scenarios that involve coordinating multiple applications, integrating diverse functionalities across apps, or executing extended sequences of operations within a single application. Recently, several long-horizon benchmarks have emerged across different platforms, including ColorBench~\citep{colorbench} for mobile, VeriWeb~\citep{veriweb} and RealWebAssist~\citep{realwebassist} for web. However, these long-horizon benchmarks are either offline benchmarks, which scale easily but cannot capture real execution dynamics (\eg, unexpected dialogs, environment state changes) and introduce offline-to-online gap, or target non-mobile platforms where partial observability and accessibility limitations are less severe. }

\rev{AW-Extend is an \textbf{online} and \textbf{long-horizon} benchmark for \textbf{mobile} agent evaluation. It runs on live Android emulators with handcrafted environment setup and automated state-based evaluation scripts to ensure realistic environment dynamics. AW-Extend tasks require over 30 steps on average, compared to 13.0 for ColorBench and 17.6 for RealWebAssist, making them substantially more challenging for context management. AW-Extend utilizes realistic task composition and batching (\eg,"add multiple contacts and message each one"), consistent with methodologies in ColorBench~\citep{colorbench} and VeriWeb~\citep{veriweb}. Our extended suite comprises 19 tasks covering two categories:}

\textbf{Compositional Tasks} combine multiple atomic operations into realistic workflows that require maintaining task context across different applications. These tasks are characterized by strong dependencies between subtasks, demanding that agents \textbf{retain and effectively utilize critical information} throughout the execution process. Representative examples include: processing a To-Do List from notes by completing each item sequentially (creating calendar events, adding contacts, managing expenses); batch operations such as creating multiple contacts and messaging each one; and cross-application interactions like querying note content via text message and replying accordingly. 

\textbf{Iterative Tasks} systematically scale the sub-task count from the standard $n=3$ in AndroidWorld to $n=10$ and $n=20$, testing both agent robustness and the ability to filter irrelevant information from context. Unlike compositional tasks where subtasks are interdependent, iterative tasks involve weakly-correlated operations, making them particularly effective for evaluating an agent's capacity to \textbf{suppress interference from irrelevant historical context} while maintaining consistent task execution. These tasks encompass managing expense records, calendar events, menu items, and note operations. 

We adapted AndroidWorld's environment configuration and evaluation code to create these tasks, enabling assessment through AndroidWorld's standard interface, including environment setup, automated evaluation based on emulator state, and trajectory recording.
\section{Experiments}
\subsection{Experimental Settings}

\textbf{Benchmarks.} We conduct experiments on AndroidWorld~\citep{androidworld} and AW-Extend. AndroidWorld is a comprehensive and widely used benchmark for evaluating Android GUI agents across diverse real-world mobile applications with 116 tasks. 
AW-Extend is an extension we construct with 19 additional long horizon tasks described in Section~\ref{sec:aw_extend}.

\textbf{Baselines.} We compare our method with several state-of-the-art approaches on two benchmarks. \textbf{On AndroidWorld benchmark,} as shown in Table~\ref{tab:android_world_results}, we include a wide range of methods from related work. We re-evaluated M3A method~\citep{androidworld}, for both its \textit{a11y tree} and \textit{SoM} variants, using Gemini-2.5-Pro, as the original work reported results based on GPT-4-Turbo. All other baseline scores on AndroidWorld are cited from their respective original publications. \textbf{On AW-Extend benchmark,} we select a focused set of baselines, each employing a distinct context management strategy to handle long-horizon tasks. As detailed in Table~\ref{tab:extended_results}, these include:
\begin{itemize}
    \item \textbf{M3A}~\citep{androidworld}, which generates a textual summary after each step to manage its context.
    \item \textbf{UI-TARS}~\citep{uitars}, which uses a sliding window approach, discarding the oldest screenshots in a first-in-first-out (FIFO) manner when the context exceeds a limit. We follow its official setting, retaining a maximum of five screenshots.
    \item \textbf{Mobile-Agent-v3}~\citep{mobile-agent-v3}, which employs hierarchical planning method. A planner first decomposes the main task into a list of sub-tasks, and an executor then completes them sequentially. The planner maintains the list of sub-tasks continuously during the task execution.
\end{itemize}

For baselines on AW-Extend benchmark, we use Gemini-2.5-Pro~\citep{Gemini25Pro} as the backbone model for M3A~\citep{androidworld}, UI-TARS-1.5-API for UI-TARS~\citep{uitars}, and both GUI-Owl-32B and GUI-Owl-7B for Mobile-Agent-v3~\citep{mobile-agent-v3}. \rev{We note that Mobile-Agent-v3 is designed around a specialized model (GUI-Owl) with built-in grounding capabilities, making it incompatible with direct backbone substitution to general-purpose LLMs. We evaluate it using its reported best settings to serve as a representative baseline for hierarchical planning methods.}

\textbf{Metrics.} We adapt the official AndroidWorld evaluation metric (Success Rate) to measure task completion performance. For AW-Extend, we adapt AndroidWorld's evaluation codebase to support assessment of our custom task set.

\textbf{Implementation Details.} We implement \ours using Gemini-2.5-Pro~\citep{Gemini25Pro} as the backbone planning model, and leverage UI-TARS-1.5-API~\citep{uitars} to locate the UI elements.

\subsection{Analysis of Success Rate}

\begin{table}[htbp]
    \centering
    \caption{Success Rate (\%) on AndroidWorld.}
    \label{tab:android_world_results}
        \resizebox{\columnwidth}{!}{%
            \begin{tabular}{lc}
            \toprule
            \textbf{Method} & \textbf{AndroidWorld (\%)} \\ 
            \midrule
            MobileGPT~\citep{mobilegpt} & 23.0 \\
            AutoDroid-V2~\citep{autodroidv2} & 26.0 \\
            M3A (a11y, GPT-4-Turbo)~\citep{androidworld} & 30.6 \\
            M3A (a11y, Gemini-2.5-Pro)~\citep{androidworld} & 31.0 \\
            M3A (SoM, GPT-4-Turbo)~\citep{androidworld} & 25.4\\
            M3A (SoM, Gemini-2.5-Pro)~\citep{androidworld} & 39.7\\
            GLM-4.1V-9B-Thinking~\citep{glm-4.1} & 41.7 \\
            UI-TARS (UI-TARS-7B)~\citep{uitars} & 33.0 \\
            UI-TARS (UI-TARS-1.5-API)~\citep{uitars} & 64.2 \\
            V-Droid~\citep{v-droid} & 59.5\\
            MobileUse~\citep{mobileuse} & 62.9\\
            UI-Venus~\citep{ui-venus} & 65.9\\
            Agent S3~\citep{agent-s3} & 68.1\\
            Mobile-Agent-v3 (GUI-Owl-7B)~\citep{mobile-agent-v3} & 66.4\\
            Mobile-Agent-v3 (GUI-Owl-32B)~\citep{mobile-agent-v3} & 73.3\\
            \midrule
            \textbf{\ours} & \textbf{78.0} \\ 
            \midrule
        \end{tabular}
    }
\end{table}

\begin{table}[htbp]
    \centering
    \caption{Success Rate (\%) on AW-Extend. }
    \label{tab:extended_results}
        \begin{tabular}{lc}
            \toprule
            \textbf{Method} & \textbf{AW-Extend (\%)}\\ 
            \midrule
            M3A (a11y) & 23.7\\
            M3A (SoM) & 28.9 \\
            Mobile-Agent-v3 (GUI-Owl-7B) & 26.3 \\
            Mobile-Agent-v3 (GUI-Owl-32B) & 28.9 \\
            UI-TARS (UI-TARS-1.5-API)& 36.8 \\
            \midrule
            \textbf{\ours} & \textbf{68.4} \\
            \midrule
        \end{tabular}
\end{table}

\textbf{Overall Results. } As shown in Table~\ref{tab:android_world_results}, our method achieves a success rate of 78.0\% on the AndroidWorld benchmark, surpassing the previous state-of-the-art Mobile-Agent-v3 by 4.7\%. This demonstrates significant advantages in general mobile automation tasks. On AW-Extend benchmark shown in Table~\ref{tab:extended_results}, we compare our program-guided context management approach against three representative baseline context management strategies: Summarization (M3A), sliding window (UI-TARS), and hierarchical planning (Mobile-Agent-v3). Our method achieves a success rate of 68.4\%, substantially outperforming all baseline approaches. These results demonstrate that our program-guided context management framework offers substantial advantages over existing methods, particularly for long-horizon tasks that require maintaining context effectively over extended interaction sequences.

\begin{figure}[htbp]
    \centering
    \includegraphics[width=\linewidth]{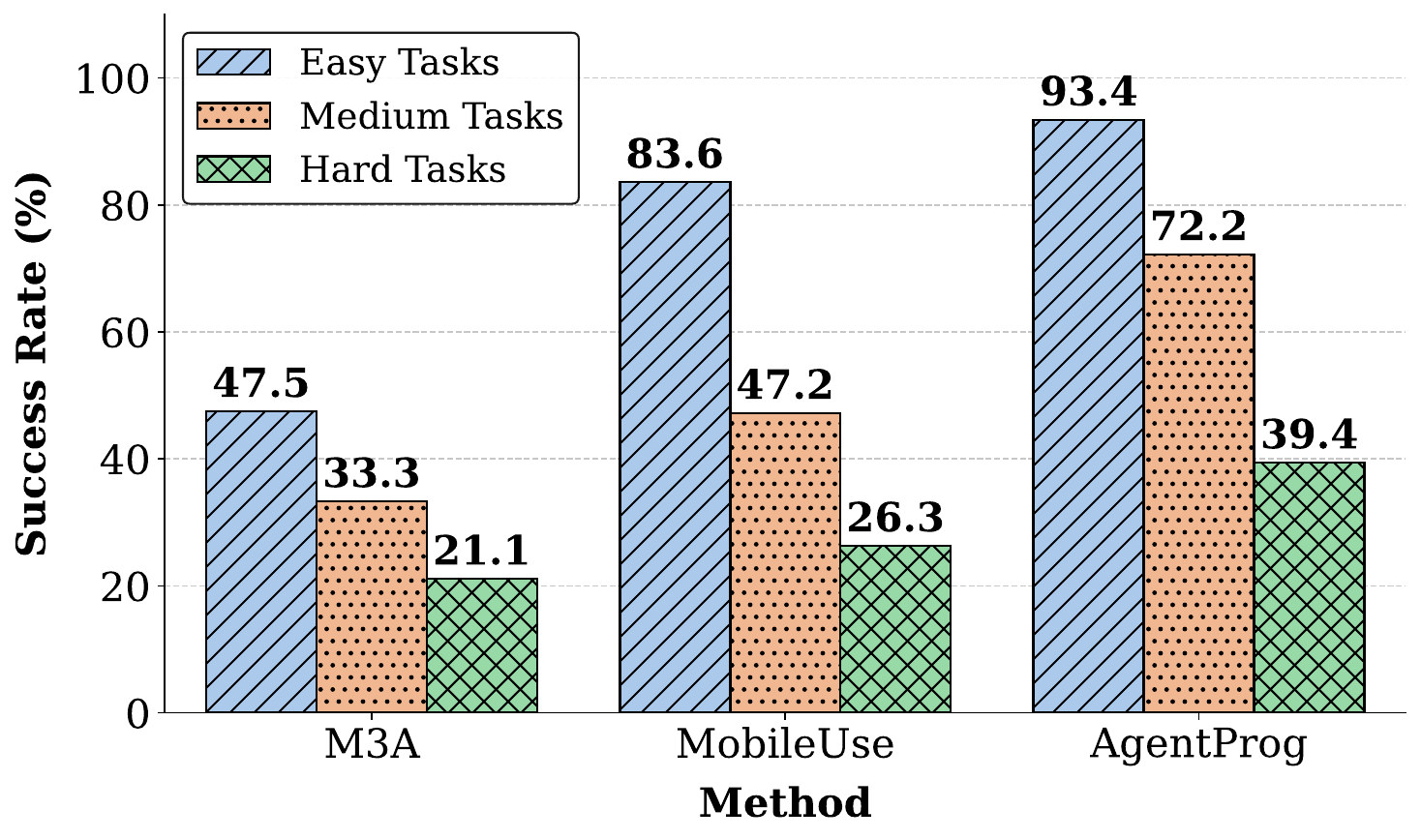}
    \caption{Success Rate (\%) across difficulty levels on AndroidWorld.}
    \label{fig:level_acc}
\end{figure}

\textbf{Performance Across Task Difficulty Levels.} Figure~\ref{fig:level_acc} presents the success rates of different methods stratified by task difficulty levels on AndroidWorld. In Figure~\ref{fig:level_acc}, M3A uses SoM as input and Gemini-2.5-Pro as backbone model. MobileUse results are reported from the original paper, while M3A and \ours results are from our evaluation. Other existing methods have not reported performance across different difficulty levels on AndroidWorld. We observe that success rates decline progressively as task difficulty increases across all methods. Notably, \ours consistently outperforms baselines at every difficulty level. Besides, the magnitude of improvement is most pronounced for medium tasks (25.0\% over MobileUse), followed by hard tasks (13.1\% improvement), with the smallest gain on easy tasks (9.8\% improvement). This pattern suggests that existing methods can handle simple tasks reasonably well, but face substantial challenges with medium and hard tasks. \ours effectively addresses this gap, particularly excelling where current methods struggle most. This demonstrates that the program-driven framework with belief state tracking in \ours is particularly effective for complex tasks that demand advanced planning capabilities, robust memory of critical information, and sophisticated environment understanding. The widening performance gap at higher difficulty levels validates that these capabilities become increasingly critical as task complexity grows.

\begin{figure}[htbp]
    \centering
    \includegraphics[width=\linewidth]{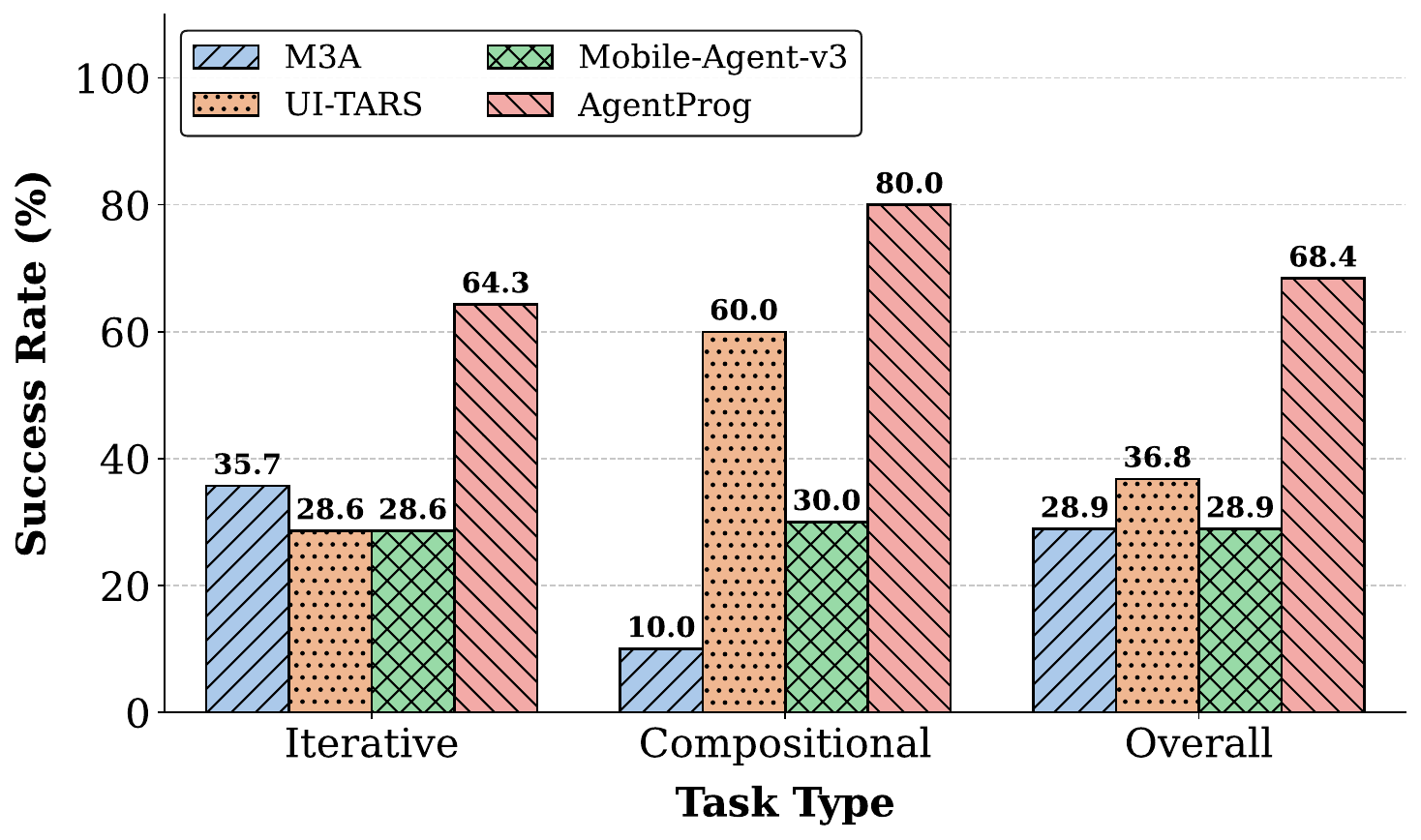}
    \caption{Success Rate (\%) across difficulty task types on AW-Extend.}
    \label{fig:type_acc}
\end{figure}

\textbf{Performance Across Different Task Types.} AW-Extend benchmark comprises two distinct task categories: iterative tasks and compositional tasks. We analyze the performance of different methods across these categories, as illustrated in Figure~\ref{fig:type_acc}. Compositional tasks combine multiple related subtasks, requiring agents to retain information from one subtask while executing another, thereby testing their ability to preserve critical memory. In contrast, iterative tasks involve executing similar operations multiple times with weak inter-task dependencies, primarily challenging the capability of agents to filter out irrelevant memory interference while maintaining robust task completion. Among only baseline methods, UI-TARS achieves the best overall performance, primarily due to its strong results on compositional tasks. This advantage stems from its sliding window approach, which avoids summarizing historical information, allowing the model to access information from previous subtasks when executing the current one. However, both UI-TARS and Mobile-Agent-v3 underperform M3A on iterative tasks. For UI-TARS, this may be attributed to the sliding window accumulating substantial irrelevant memory in the context, which interferes with model performance. The more complex architecture of Mobile-Agent-v3 compared to M3A introduces greater instability in context management during task execution, reducing robustness for iterative operations. M3A outperforms other baselines on iterative tasks, likely because its summary-based context management offers a simpler, more stable architecture that maintains lower failure rates when executing similar operations repeatedly. \ours surpasses all baseline methods, demonstrating significant advantages on both iterative and compositional tasks. The program-guided context management approach enables robust completion of both task types: it effectively preserves important information through program variables while eliminating redundant and irrelevant memory by parsing program structure. This dual capability allows \ours to excel where existing methods face trade-offs between memory retention and interference reduction.

\textbf{Impact of Global Belief State.} We further investigate the contribution of the global belief state component by comparing \ours with a variant that removes this module. The results reveal substantial performance improvements achieved by incorporating belief state tracking: the success rate improves from 53.9\% to 78.0\% on AndroidWorld (increased by 24.1\%), and from 35.1\% to 68.4\% on AW-Extend (increased by 33.3\%). These significant gains underscore the critical role of global belief state in long-horizon tasks. The global belief state mechanism enhances the agent's understanding of environmental states, enabling it to detect and recover from unexpected situations immediately. This capability is particularly valuable for maintaining execution stability, which becomes increasingly crucial as task horizons extend. The more pronounced performance improvement on AW-Extend further validates that belief state tracking becomes even more essential for complex, multi-step tasks. By maintaining a structured representation of the current environment state, the global belief state allows~\ours~to make more informed decisions, reduce cascading errors, and ultimately achieve more robust task completion across diverse scenarios.

\rev{\textbf{Impact of Execution Tree.} Removing Execution Tree causes severe degradation on AW-Extend, dropping from 68.4\% to 39.5\%. On AndroidWorld, performance also decreases substantially from 78.0\% to 61.6\%. Without the Execution Tree, the agent retains all historical steps in a flat sequence, causing irrelevant context from completed loop iterations and non-taken branches to accumulate rapidly. This is particularly disadvantageous for iterative tasks in AW-Extend, where dozens of similar subtask histories overwhelm the agent's context and interfere with decision-making. The result confirms that control flow-based context pruning is indispensable for long-horizon tasks.}

\rev{\textbf{Impact of Explicit Variables.} Removing Explicit Variables leads to a performance drop from 78.0\% to 64.2\% on AndroidWorld and from 68.4\% to 50.0\% on AW-Extend. Without declared variables, the agent loses its ``anchors'' for preserving task-critical data (\eg, a list of to-do items or extracted contact information) across STP execution phase. The impact is smaller than removing the Execution Tree because the low-level Python code generation still allows the LLM some autonomy to define and track variables implicitly.}

\rev{\textbf{STP Error Analysis.} To assess the robustness of Semantic Task Program generation, we analyze STP errors across all tasks in AndroidWorld. We define an STP error as a case where the generated program contains incorrect task decomposition, missing steps, or invalid assumptions about app workflows that cannot be recovered during execution. Out of all generated STPs, only 2.5\% contain such errors, indicating that the STP generation process is highly reliable. We attribute this robustness to STP specifies core business logic (\eg, CRUD actions) without over-decomposing tasks into low-level operations, delegating the handling of unseen states and implicit prerequisites to the execution phase. Moreover, our architecture is compatible with existing re-planning strategies~\citep{reflexion, adaplanner}: Global Belief State mechanism can detect misalignment between STP assumptions and the dynamic environment at runtime, so it is possible for \ours to identify the failure cause and provide feedback for adaptive STP correction. Integration with re-planning modules remains a promising direction for future investigation.}

\rev{\textbf{A Runtime Recovery Example of Global Belief State.} Global Belief State helps detect and mitigate situations where the high-level STP makes incorrect assumptions about unseen app states. We illustrate this with the \textit{BrowserMaze} task from AndroidWorld, where the agent must open a maze HTML file in Chrome and navigate to the bottom-right goal. The generated STP assumes an obstacle-free grid and plans a straightforward strategy: ``repeatedly click \emph{down} until no further movement, then repeatedly click \emph{right}.'' However, the maze contains obstacles that block this linear path. During execution, GBS tracks the agent's position and detects a Belief-Reality Gap: the expected position after a \emph{right} click does not change, indicating a blocked cell. Rather than continuing the failing strategy, GBS updates its belief to reflect the obstacle and triggers an adaptive rerouting: the agent alternates between \emph{down} and \emph{right} clicks and successfully reaches the bottom-right goal. This case demonstrates how GBS enables robust execution by detecting misalignment between STP assumptions and the actual environment at runtime. Further details and visualizations are provided in Appendix~\ref{sec:gbs_case}.}

\subsection{Analysis of Latency and Cost}

\begin{table*}[htbp]
\centering
\caption{Per-Task Token Consumption and Latency for Successful Attempts on AW-Extend. \textnormal{``Static Prefix'' represents the static prefix prompt tokens that remain constant across invocations, ``Dynamic'' represents the prompt tokens that continuously change during task execution, and ``Output'' represents the total number of tokens generated during task execution.}}  
\label{tab:token_cost}
\begin{tabular}{llrrrr}
\toprule
\textbf{Agent} & \textbf{Task Set} & \textbf{Static Prefix (k)} & \textbf{Dynamic (k)} & \textbf{Output (k)} & \textbf{Latency (s)} \\
\midrule
\multirow{3}{*}{UI-TARS} & Overall               & 8.1 & 315.5 & 3.2 & 312 \\
                         & Iterative  & 8.5 & 329.6 & 3.3 & 344 \\
                         & Compositional & 7.6 & 294.3 & 3.0 & 263 \\
\midrule
\multirow{3}{*}{Mobile-Agent-v3} & Overall               & 7.9 & 809.4 & 22.1 & 1604 \\
                                 & Iterative  & 8.9 & 917.6 & 26.1 & 2033 \\
                                 & Compositional & 5.8 & 592.9 & 14.2 &  746 \\
\midrule
\multirow{3}{*}{\ours} & Overall               & 1026.4 & 301.3 & 179.7 & 2662 \\
                           & Iterative  & 1275.3 & 370.3 & 225.1 & 2910 \\
                           & Compositional &  777.4 & 232.4 & 134.3 & 2164 \\
\bottomrule
\end{tabular}
\end{table*}

\textbf{Token Consumption and Latency.} As shown in Table~\ref{tab:token_cost}, we compare the token consumption and latency per task across \ours and two baseline methods: Mobile-Agent-v3 and UI-TARS, which outperform other baselines in AndroidWorld and AW-Extend respectively. First, we analyze token consumption by distinguishing between static prefix and dynamic prompt tokens since \ours is based on prompt engineering with substantial system prompts, a large portion of the prefix can be cached to reduce cost and improve efficiency. This distinction provides meaningful insights into the true efficiency of different approaches. Second, we compare only tasks where all three methods succeeded, as failed tasks exhibit inconsistent token patterns—some fail due to premature termination, leading to low token counts,  while others fail after excessive iterative trial-and-error operations, leading to high token counts. Including failed tasks would obscure the true efficiency characteristics of each method. 

Table~\ref{tab:token_cost} shows that UI-TARS demonstrates the lowest latency and minimal token consumption because its backbone model trained on extensive data with short static prompts and performs no additional context processing (\eg, summarization), resulting in minimal output tokens and the fastest execution speed. In contrast, Mobile-Agent-v3, while also using a trained model with relatively short static prefix (53 tokens per call), requires more tokens than UI-TARS due to its summarization and reflection process. Since \ours is implemented through prompt engineering, it contains a substantial static prompt prefix (12.5k tokens per call), resulting in a much higher proportion of cached tokens in the input, which reduces the cost of input tokens. Notably, \ours achieves significantly lower dynamic token counts compared to Mobile-Agent-v3, with performance comparable to UI-TARS. This demonstrates the effectiveness of~\ours's context management approach over hierarchical planning method in maintaining compact context. However, \ours generates significantly more output tokens compared to both baselines, as it produces extensive intermediate content for context management including belief state, and many operations involve program counter update and variable manipulations. In terms of latency, \ours exhibits longer execution time than Mobile-Agent-v3, primarily attributable to the increased output token generation. All three methods exhibit consistent trends across task categories: iterative tasks involve numerous repeated subtasks and thus require higher token consumption, while compositional tasks contain fewer combined subtasks and consequently demand fewer tokens.

\begin{figure}[htbp]
    \centering
    \includegraphics[width=1\linewidth]{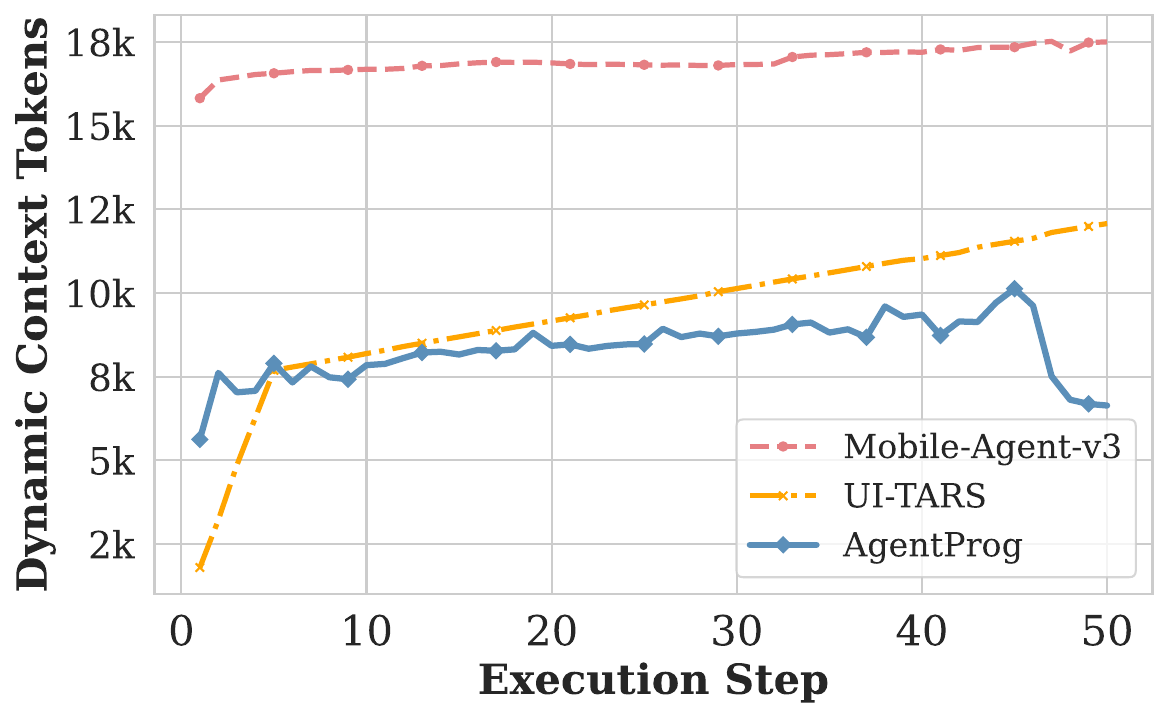}
    \caption{Dynamic context tokens in 50 steps.}
    \label{fig:memory_dynamic}
\end{figure}

\textbf{Context Tokens Over Steps.} As shown in Figure~\ref{fig:memory_dynamic}, we analyze the dynamic context growth patterns of different methods over extended task execution in 50 steps. We compute the average dynamic tokens at each step across all agent trajectories on the AW-Extend benchmark. UI-TARS performs minimal processing of historical information, only discarding excess screenshots when necessary. Consequently, a notable drawback of this approach is the continuous growth of context length over time. UI-TARS exhibits rapid dynamic context growth in the first 5 steps, primarily because each step adds a new screenshot to the context. From step 6 onwards, the growth rate slows as UI-TARS begins discarding the oldest screenshot while adding a new one, with context expansion mainly driven by the textual portion of historical information. 

Mobile-Agent-v3 also achieves relatively slow dynamic token growth thanks to its hierarchical planning mechanism, which decomposes complex tasks into subtasks and progressively removes information during execution. However, Mobile-Agent-v3 maintains dynamic tokens at a consistently high level (approximately 17k tokens) throughout task execution. In contrast, \ours consistently maintains dynamic tokens at a lower level (approximately 9k tokens), demonstrating advantages over both baselines. During 50 steps, the dynamic context token count in \ours remains nearly constant, indicating the effectiveness of our program-guided management in maintaining a compact representation of context. This stability is achieved through our approach's ability to represent critical information in program variables and global belief state, while systematically pruning redundant and irrelevant historical details by parsing program structure. \ours scales efficiently to long-horizon tasks, ensuring consistent performance even as task complexity and duration increase, while requiring much fewer dynamic tokens than existing methods.
\begin{table}[htbp]
\centering
\caption{\textbf{Per-task average steps on AW-Extend.}}
\label{tab:step_comparison}
\resizebox{\columnwidth}{!}{%
\begin{tabular}{lccc}
\toprule
\textbf{Task Set} & \textbf{\ours} & \textbf{Mobile-Agent-v3} & \textbf{UI-TARS}\\
\midrule
\multicolumn{4}{l}{\textit{All Attempts}} \\
\midrule
Overall & 129.6 & 134.0 & 225.5\\
Iterative & 138.5 & 169.1 & 224.9\\
Compositional & 104.6 & 34.0 & 227.0\\
\midrule
\multicolumn{4}{l}{\textit{Successful Attempts Only}} \\
\midrule
Overall  & 37.7 & 51.2 & 43.6\\
Iterative & 36.8 & 57.8 & 50.0\\
Compositional & 39.5 & 38.0 & 34.0\\
\bottomrule
\end{tabular}%
}
\end{table}

\textbf{Average Steps per Task.} Table~\ref{tab:step_comparison} presents the average number of steps required per task for \ours and baselines on the AW-Extend dataset. Step counts should be interpreted with caution, as different methods require varying numbers of model queries per step: Mobile-Agent-v3 queries the language model three times per step and UI-TARS queries once per step while \ours uses two queries per step. The trends in step counts largely match those observed in token consumption as shown in Table~\ref{tab:token_cost}. When considering all attempts, the compositional task results show an anomalous pattern where Mobile-Agent-v3 uses very few steps, attributable to premature task termination due to forgotten subtasks. When examining only successful attempts, \ours requires fewer steps than baselines across the overall tasks and iterative tasks, with comparable performance on composition tasks. This demonstrates that \ours achieves better performance while maintaining step efficiency. The higher step counts in ``All Attempts'' group compared to ``Successful Attempts Only'' group can be attributed to two factors: first, tasks requiring fewer steps are inherently easier to complete successfully; second, failed attempts in ``All Attempts'' involve substantial iterative trial-and-error operations that inflate the step count.

\section{Discussion}

\textbf{\rev{Accuracy-Latency Trade-off.}} \rev{As shown in Table~\ref{tab:token_cost}, \ours incurs higher latency and cost compared to baseline methods. This overhead stems primarily from the STP execution, including Program Counter Update, Action Generation, and Global Belief State maintenance. However, we argue that \ours primarily targets complex, long-horizon tasks suited for background execution, where users are far more sensitive to accuracy than latency, as fast agents become impractical if they require frequent human intervention for failures. We anticipate future model-level optimizations (\eg, faster inference speeds, distilled models) and system optimizations such as merging the Program Counter Update and Action Generation modes into a single LLM call where possible.}

\textbf{\rev{On-Device Deployment.}} \rev{Currently, \ours relies on cloud-based LLMs, which introduce network latency and privacy concerns. A promising future direction is deploying \ours with on-device Small Language Models (SLMs) via model distillation~\citep{autodroidv2}. Our program-guided context management reduces the input dynamic context size and the static context can be eliminated entirely through distillation, which is compatible with the limited context lengths of on-device SLMs. Besides, \ours supports a hybrid deployment strategy. STP Generation phase requiring stronger reasoning capabilities and more knowledge can remain on the cloud, while the STP Execution phase can be distilled into an on-device SLM, offering a flexible trade-off among latency, privacy, and accuracy.}

\rev{\textbf{Generalization to Broader Long-Horizon Agents.} Although \ours is evaluated on mobile GUI tasks, the program-guided context management approach can be applied to other long-horizon agent settings,  including web navigation, desktop automation, and tool-use environments. We leave the empirical validation of \ours on non-mobile settings to future work.}
\section{Conclusion}

In this work, we introduce \ours to address the context bottleneck in long-horizon tasks. 
By reframing interaction history into a structured {Semantic Task Program}, \ours enables principled context pruning and critical variable retention. 
Additionally, our {Global Belief State} mechanism maintains an active mental model to handle partial observability and environmental dynamics. 
Experiments on AndroidWorld and our AW-Extend benchmark demonstrate that \ours significantly outperforms state-of-the-art baselines. 
Ultimately, our work establishes a scalable foundation for robust agent deployment in complex real-world scenarios.
\section*{Acknowledgements}
This research was supported in part by the National Natural Science Foundation of China under Grant No. 62432004 and 62272261, the Fundamental and Interdisciplinary Disciplines Breakthrough Plan of the Ministry of Education of China under Grant No. JYB2025XD\\XM122, Wuxi Research Institute of Applied Technologies, Tsinghua University under Grant 20242001120, Tsinghua University (AIR)–AsiaInfo Technologies (China) Inc. Joint Research Center and a grant from the Guoqiang Institute, Tsinghua University.

\balance
\bibliographystyle{ACM-Reference-Format}
\bibliography{reference}

\appendix
\newcommand{\sys}{Semantic Task Program\xspace}
\lstdefinestyle{myprompt}{
    breaklines=true,          
    basicstyle=\ttfamily\scriptsize, 
    keywordstyle={},          
    stringstyle={},
    commentstyle={},
    identifierstyle={},
    language={}               
}
\section{Syntax of \sys}
\label{sec:dsl}

The syntax of \sys (STP) is designed to resolve the conflict between the need for structural rigor in workflows and the inherent ambiguity of agent tasks. While \sys adopts a natural-language-style appearance to ensure expressiveness, it is fundamentally a structured language. It imposes a rigid skeleton of control flow (loops, conditionals, functions) to maintain logical consistency, while allowing the atomic instructions within that skeleton to remain "soft" and flexible.

This design philosophy differs from traditional coding, which demands precision everywhere, and prompt engineering, which lacks structural constraints. \sys allows developers to be precise about procedure (\eg, "iterate exactly 10 times") while being fuzzy about implementation (\eg, "extract the sentiment"). This section outlines how \sys implements this hybrid approach.

\subsection{Basic Statements and Comments}
\sys adopts an \textbf{intent-based programming style}. Instructions are written as imperative or declarative sentences that prioritize conveying the user's goal over adhering to strict syntactic rules.

\begin{lstlisting}
# This is a comment, used for human-readable annotations.
tell user "Hello, world!"  # Inline comments are also supported.

# The following statements are also valid:
send "Hello, world!" to the user
notify user with message "Hello, world!"
send a greeting message to user
\end{lstlisting}

Comments are ignored by \ours and serve only to document the code. Statements like \texttt{tell user "..."} are direct commands that are translated into actions. For familiarity, \sys also permits Python-style syntax where appropriate (\eg, \texttt{print("Hello, world!")}), which the \ours processes as an equivalent intent rather than executing directly.

\subsection{Variables and Data Management}
In agentic workflows, data is often unstructured (\eg, natural text, web content) and does not fit neatly into the rigid types of C++ or Java. \sys addresses this by treating variables as named containers for information, explicitly demarcated by curly braces \texttt{\{\}} (\eg, \texttt{\{userName\}}). This notation serves a dual purpose: it provides a clear anchor for the \ours to track state within a fuzzy instruction, and it allows the language to handle unstructured data without requiring complex schema definitions upfront.

\textbf{Declaration and Assignment:} Variables are declared implicitly upon their first assignment. \sys supports multiple natural phrasings for assignment, treating them as synonyms.

\begin{lstlisting}
set variable {userName} to "Alice"
store 100 into {initialScore}
set {userCount} to 0 # Python-style assignment is also valid.
\end{lstlisting}

\textbf{Using Variables:} To reference a variable's value, simply include its braced name within a statement. The \ours substitutes it with its current value during execution.

\begin{lstlisting}
tell user "Welcome, {userName}! Your score is {initialScore}."
calculate {initialScore} + 50, record as {finalScore}
\end{lstlisting}

\textbf{Dynamic Type System:} \sys employs a dynamic type system where users are not required to declare types explicitly. The \ours infers and manages types—such as Text, Number, Boolean, List, and Object—based on context and value. This approach aligns with the needs of both novice programmers and LLMs, which may struggle with strict static typing. While types are inferred, users can provide hints to guide the interpreter, for instance, when interacting with an AI model.

\begin{itemize}
    \item \textbf{List:} An ordered collection.
\begin{lstlisting}
record list "apples", "bananas", "cherries" as {fruitBasket}
\end{lstlisting}
    \item \textbf{Object:} A key-value store, akin to a dictionary or JSON object. Properties can be accessed using dot notation (\texttt{\{product.price\}}) or possessive phrasing (\texttt{\{myCar's color\}}).
\begin{lstlisting}
create an object {product} with "name" as "Laptop" and "price" as 1200
tell user "{product.name} is available for {product.price}."
\end{lstlisting}
    \item \textbf{Table:} A structured list of objects, ideal for representing tabular data.
\begin{lstlisting}
read table "sales_data.csv" as {salesReport}
\end{lstlisting}
\end{itemize}

Crucially, the \ours handles potential type mismatches intelligently. Instead of crashing, it attempts a reasonable type coercion (\eg, converting the text \texttt{"123"} to a number) or reports a descriptive, human-understandable issue.

\subsection{Control Flow}
Control flow structures in \sys use natural language keywords and rely on indentation to define code blocks, similar to Python. This maintains readability while avoiding complex bracketing.

\textbf{Conditional Statements:} \texttt{if}, \texttt{else}, \texttt{else if} structures evaluate conditions expressed in natural language or with standard operators.

\begin{lstlisting}
ask user "Enter your age:", get response as {ageInput}
convert {ageInput} to number, as {userAge}
if {userAge} < 18:
    tell user "You are a minor."
else, if {userAge} is between 18 and 65: # Natural language condition allowed
    tell user "You are an adult."
else:
    tell user "You are a senior citizen."
\end{lstlisting}

\textbf{Loops:} \sys supports various loop constructs, each initiated with flexible, descriptive phrasing.
\begin{itemize}
    \item \textbf{Fixed Repetitions:}
\begin{lstlisting}
repeat 5 times:
    # {loop.iteration} is a special variable (1-indexed)
    tell user "This is message number {loop.iteration}"
\end{lstlisting}
    
    \item \textbf{Iterating over Collections:}
\begin{lstlisting}
iterate through each item in {fruitBasket} as {fruit}:
    tell user "Found fruit: {fruit}"
\end{lstlisting}

    \item \textbf{Conditional Loops:}
\begin{lstlisting}
while {systemStatus} is "active":
    check for new messages
    wait 10 seconds
\end{lstlisting}
\end{itemize}

\subsection{Functions}
To promote modularity and code reuse, \sys allows users to define reusable blocks of logic as \textbf{functions}.

\textbf{Defining a Function:} A function is defined with a name, a list of expected inputs (parameters), and a body of instructions.

\begin{lstlisting}
define function named "calculateArea":
    function inputs: {length} (number), {width} (number)
    calculate {length} * {width}, record as {area}
    function returns {area}
\end{lstlisting}

\textbf{Executing a Function:} Functions are called by name, with arguments passed using a clear key-value format.

\begin{lstlisting}
execute function "calculateArea", with {length} as 10 and {width} as 5, save result as {roomArea}
tell user "The room area is {roomArea} square units."
\end{lstlisting}

\subsection{Interactions with External Tools}
A cornerstone of \sys is its native ability to interact with the broader digital environment, including AI models and external tools, through high-level, declarative commands.

\textbf{Tool Use (File System, Browser):} \sys provides a unified, high-level interface for common tool interactions. This design allows users to specify *what* they want to achieve, leaving the *how* to the \ours.

\begin{lstlisting}
# File System
read text from file "config.txt" as {configContent}
save {summary} to file "summary.txt"

# Web Browser Automation
open URL "https://en.wikipedia.org"
search "Hello World"
wait for page to load
get current page content as {wikiPageContent}
\end{lstlisting}

The \ours's context-aware nature is critical here. It resolves ambiguous references like \texttt{"search input"} or \texttt{"Search button"} by analyzing the current state of the web page (its DOM structure and visual layout), making automation scripts more robust and less brittle.

\section{Runtime Recovery Example of Global Belief State}
\label{sec:gbs_case}

\begin{figure*}[!htb]
    \centering
    \includegraphics[width=0.8\textwidth]{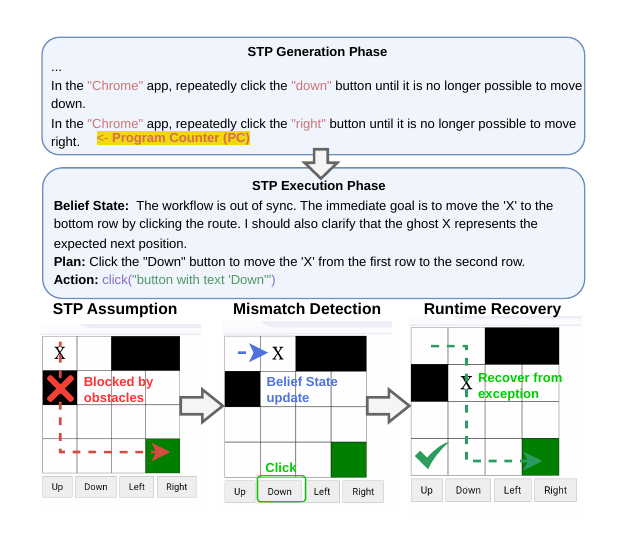}
    \caption{\rev{A runtime recovery example of global belief state.}}
    \label{fig:gbs_example}
\end{figure*}

\rev{Figure \ref{fig:gbs_example} demonstrates a runtime recovery example of Global Belief State.}

\section{Tasks in AW-Extend}
\label{sec:aw_extend_tasks}

\rev{Table \ref{tab:aw_extend_tasks} shows the task descriptions of all tasks in AW-Extend.}

\begin{table*}[htbp]
\centering
\caption{\rev{Task descriptions in AW-Extend. ``Long'' denotes tasks with $n{=}10$ subtasks and ``SuperLong'' denotes $n{=}20$.}}
\label{tab:aw_extend_tasks}
\rev{
\begin{tabular}{l p{10cm}}
\toprule
\textbf{Task Name} & \textbf{Instruction Template} \\
\midrule
ContactsAddContactAndSms & Create a new contact for \{name\}. Their number is \{number\}. Then, send `hello, \{name\}' to \{number\} via SMS using Simple SMS Messenger. \\
ContactsAddMultipleContactsAndSms & For the following persons: \{contacts\}, add them as new contacts, and then use Simple SMS Messenger to send each of them a `hello, [Name]' message. \\
MarkorFetchMultipleNotesAndSms & Retrieve the file name they sent you via Simple SMS Messenger, open it in Markor, and then send its contents back to that number for each number in \{numbers\}. \\
MarkorFetchNoteAndSms & Obtain the file name sent to you by \{number\} in Simple SMS Messenger, open the file in Markor, and send its contents back to \{number\}. \\
MarkorTodoList & Open my Markor note \{file\_name\} and complete all the tasks on the list. \\
SimpleCalendarDeleteEventsLong & In Simple Calendar Pro, delete all the calendar events on \{year\}-\{month\}-\{day\}. \\
SimpleCalendarDeleteEventsOnRelativeDayLong & In Simple Calendar Pro, delete all events scheduled for this \{day\_of\_week\}. \\
SimpleCalendarDeleteEventsSuperLong & In Simple Calendar Pro, delete all the calendar events on \{year\}-\{month\}-\{day\}. \\
SimpleCalendarDeleteEventsOnRelativeDaySuperLong & In Simple Calendar Pro, delete all events scheduled for this \{day\_of\_week\}. \\
ExpenseAddMultipleLong & Add the following expenses into the Pro Expense: \{expenses\}. \\
ExpenseAddMultipleSuperLong & Add the following expenses into the Pro Expense: \{expenses\}. \\
ExpenseDeleteMultipleSuperLong & Delete the following expenses from Arduia Pro Expense: \{expenses\}. \\
ExpenseDeleteMultipleLong2 & Delete the following expenses from Arduia Pro Expense: \{expenses\}. \\
ExpenseDeleteMultipleSuperLong2 & Delete the following expenses from Arduia Pro Expense: \{expenses\}. \\
RecipeAddMultipleRecipesLong & Add the following recipes into the Broccoli app: \{recipes\}. \\
RecipeAddMultipleRecipesSuperLong & Add the following recipes into the Broccoli app: \{recipes\}. \\
RecipeDeleteMultipleRecipesLong & Delete the following recipes from Broccoli app: \{titles\}. \\
RecipeDeleteMultipleRecipesSuperLong & Delete the following recipes from Broccoli app: \{titles\}. \\
MarkorMergeNotesLong & Merge the contents of Markor notes \{file1\_name\}, \{file2\_name\} and \{file3\_name\} into a new note named \{new\_file\_name\}. Add a new line between each. \\
\bottomrule
\end{tabular}
}
\end{table*}
\section{End-to-End Pipeline Walkthrough}
\label{sec:walkthrough}

\rev{We provide a concrete walkthrough of a representative AW-Extend task that combines explicit variables, loop control flow, cross-app execution, and persistent belief tracking. The user instruction is:}

\rev{\textit{For the following persons: Name: Hana Ferreira, Number: +10662908339, Name: Sophie Martin, Number: +18723713947, Name: Olivia Alves, Number: +16278036185,add them as new contacts, and then use Simple SMS Messenger to send each of them a  'hello, [Name]' message, where [Name] is the name of each contact.}}

\rev{\textbf{STP Generation Phase.} The high-level user request is first converted into the following STP:}

\begin{lstlisting}[style=myprompt]
# Create a list of objects to store the contact information
Create a list of objects named {contact_list}, with the following items:
    An object with "name" set to "Hana Ferreira" and "number" set to "+10662908339"
    An object with "name" set to "Sophie Martin" and "number" set to "+18723713947"
    An object with "name" set to "Olivia Alves" and "number" set to "+16278036185"

# Iterate through each contact to add them and send a message
Iterate through each item in {contact_list}, recorded as {current_contact}:
    # Add the person as a new contact
    In the `Contacts` app, add a new contact with name {current_contact.name} and phone number {current_contact.number}, and save it.

    # Send a personalized SMS message
    In the `Simple SMS Messenger` app, send a message to {current_contact.number} with the content "hello, {current_contact.name}".
\end{lstlisting}

\rev{\textbf{STP Execution Phase.} In STP Execution Phase, \ours alternates between two modes to executes STP.}

\rev{\textbf{Representative prompt instances.} At execution time, the LLM receives a fixed system prefix plus a dynamic context. Below we show the \emph{dynamic} prompt content for some non-trivial steps. This is the information the model uses to ground the current STP instruction into low-level actions.}

\rev{\textbf{Action Generation prompt instance.} Here is the prompt of a single step during the execution of STP:}

\begin{lstlisting}[style=myprompt]
Mode: Action Generation

STP Context:

```
Create a list of objects named {contact_list}, with the following items:
    - An object with "name" set to "Hana Ferreira" and "number" set to "+10662908339"
    - An object with "name" set to "Sophie Martin" and "number" set to "+18723713947"
    - An object with "name" set to "Olivia Alves" and "number" set to "+16278036185"
Iterate through each item in {contact_list}, recorded as {current_contact}:
    In the `Contacts` app, add a new contact with name {current_contact.name} and phone number {current_contact.number}, and save it.  # <-- current step
    In the `Simple SMS Messenger` app, send a message to {current_contact.number} with the content "hello, {current_contact.name}". 
```

Current Line: In the `Contacts` app, add a new contact with name {current_contact.name} and phone number {current_contact.number}, and save it.

Low Level Code History:
# Workflow step: Iterate through each item in {contact_list}, recorded as {current_contact}:
contact_list_iterator = iter(contact_list)
current_contact = next(contact_list_iterator)
# Workflow step: In the `Contacts` app, add a new contact with name {current_contact.name} and phone number {current_contact.number}, and save it.

Variables:
global variables:
llm: <FoundationModel>
mobile: <MobileAPI(config=MobileAPIConfig(locator='ui_tars', device_serial_id='emulator-5564', llm=<agentprog.all_ut...)>
current_screenshot: <image>
contact_list (list, len=3): [{'name': 'Hana Ferreira', 'number': '+10662908339'}, {'name': 'Sophie Martin', 'number': '+18723713947'}, {'name': 'Olivia Alves', 'number': '+16278036185'}]
is_exec_success: True
contact_list_iterator: <list_iterator>
current_contact (dict, len=2): {'name': 'Sophie Martin', 'number': '+18723713947'}

Global Belief State:
- I am at the home screen of the mobile device.
- The `contact_list` variable is populated with three contacts.
- The workflow has started iterating through the `contact_list`.
- The `current_contact` variable now holds the information for "Sophie Martin".
- The next step is to open the `Contacts` app to add "Sophie Martin" as a new contact.
\end{lstlisting}

\rev{\ours tries to output an action to operate the device, and update the belief state:}

\begin{lstlisting}[style=myprompt]
Updated Belief State:
- I am inside the `Contacts` app, which currently shows no contacts.
- The `contact_list` variable is populated with three contacts.
- The workflow has started iterating through the `contact_list`.
- The `current_contact` variable now holds the information for "Sophie Martin".
- The `Contacts` app is open. To proceed, I need to find and click the button to add a new contact.
- A button to add a new contact, possibly labeled "Create new contact" or represented by a "+" icon, should be present on this screen, even if it's not clearly visible in the current description.

Action: 
mobile.start_app(app_name="Contacts")

\end{lstlisting}

\rev{\textbf{PC Update prompt instance.} After Action Generation, \ours is about to update the Program Counter:}

\begin{lstlisting}[style=myprompt]
Mode: Program Counter Update

STP Context:

```
Create a list of objects named {contact_list}, with the following items:
    - An object with "name" set to "Hana Ferreira" and "number" set to "+10662908339"
    - An object with "name" set to "Sophie Martin" and "number" set to "+18723713947"
    - An object with "name" set to "Olivia Alves" and "number" set to "+16278036185"
Iterate through each item in {contact_list}, recorded as {current_contact}:
    In the `Contacts` app, add a new contact with name {current_contact.name} and phone number {current_contact.number}, and save it.  # <-- current step
    In the `Simple SMS Messenger` app, send a message to {current_contact.number} with the content "hello, {current_contact.name}". 
```

Current Line: In the `Contacts` app, add a new contact with name {current_contact.name} and phone number {current_contact.number}, and save it.

Low Level Code History:
# Workflow step: Iterate through each item in {contact_list}, recorded as {current_contact}:
current_contact = next(contact_list_iterator)
# Workflow step: In the `Contacts` app, add a new contact with name {current_contact.name} and phone number {current_contact.number}, and save it.
mobile.start_app(app_name="Contacts")

Variables:
global variables:
llm: <FoundationModel>
mobile: <MobileAPI(config=MobileAPIConfig(locator='ui_tars', device_serial_id='emulator-5564', llm=<agentprog.all_ut...)>
current_screenshot: <image>
contact_list (list, len=3): [{'name': 'Hana Ferreira', 'number': '+10662908339'}, {'name': 'Sophie Martin', 'number': '+18723713947'}, {'name': 'Olivia Alves', 'number': '+16278036185'}]
is_exec_success: True
contact_list_iterator: <list_iterator>
current_contact (dict, len=2): {'name': 'Sophie Martin', 'number': '+18723713947'}

Global Belief State:
- I am inside the `Contacts` app, which currently shows no contacts.
- The `contact_list` variable is populated with three contacts.
- The workflow has started iterating through the `contact_list`.
- The `current_contact` variable now holds the information for "Sophie Martin".
- The `Contacts` app is open. To proceed, I need to find and click the button to add a new contact.
- A button to add a new contact, possibly labeled "Create new contact" or represented by a "+" icon, should be present on this screen, even if it's not clearly visible in the current description.
\end{lstlisting}

\rev{Note that the low-level action histories of adding other contacts are pruned, as the steps are similar across iterations. At each PC Update, \ours selects a Program Counter Operation from four options: \textit{hold} (remain at the current step), \textit{continue} (jump to the next step), \textit{break} (exit the loop or jump to the else-branch), or \textit{return} (exit the current function):}

\begin{lstlisting}[style=myprompt]
Updated Belief State:
- I am inside the `Contacts` app, which currently shows no contacts.
- The `contact_list` variable is populated with three contacts.
- The workflow has started iterating through the `contact_list`.
- The `current_contact` variable now holds the information for "Sophie Martin".
- The `Contacts` app is open. To proceed, I need to find and click the button to add a new contact.
- A button to add a new contact, possibly labeled "Create new contact" or represented by a "+" icon, should be present on this
screen, even if it's not clearly visible in the current description. 

Action: 
WorkflowProgramCounterOperation.HOLD
\end{lstlisting}

\rev{\textbf{PC Update prompt instance for breaking the loop.} When all contacts are added, \ours tries to get next item from the iterator and receives a \textit{StopIteration} error. Then, \ours will break the loop:}

\begin{lstlisting}[style=myprompt]
Mode: Program Counter Update

STP Context:

```
Create a list of objects named {contact_list}, with the following items:
    - An object with "name" set to "Hana Ferreira" and "number" set to "+10662908339"
    - An object with "name" set to "Sophie Martin" and "number" set to "+18723713947"
    - An object with "name" set to "Olivia Alves" and "number" set to "+16278036185"
Iterate through each item in {contact_list}, recorded as {current_contact}:  # <-- current step
    In the `Contacts` app, add a new contact with name {current_contact.name} and phone number {current_contact.number}, and save it.
    In the `Simple SMS Messenger` app, send a message to {current_contact.number} with the content "hello, {current_contact.name}". 
```

Current Line: Iterate through each item in {contact_list}, recorded as {current_contact}:

Low Level Code History:
# Workflow step: Iterate through each item in {contact_list}, recorded as {current_contact}:
current_contact = next(contact_list_iterator)
# Execute Result: '[StopIteration]: \n-> current_contact = next(contact_list_iterator)'

Variables:
global variables:
llm: <FoundationModel>
mobile: <MobileAPI(config=MobileAPIConfig(locator='ui_tars', device_serial_id='emulator-5564', llm=<agentprog.all_ut...)>
current_screenshot: <image>
contact_list_iterator: <list_iterator>
current_contact (dict, len=2): {'name': 'Olivia Alves', 'number': '+16278036185'}
first_name: 'Olivia'
last_name: 'Alves'
error_info: '[StopIteration]: \n-> current_contact = next(contact_list_iterator)'

Global Belief State:
- The contact "Sophie Martin" with number "+18723713947" has been successfully added and messaged.
- The contact "Olivia Alves" with number "+16278036185" has been successfully created.
- A message "hello, Olivia Alves" has been sent to her, and the current screen shows the SMS conversation, confirming the completion of the last iteration.
- All contacts in the list have now been processed.
\end{lstlisting}

\rev{And \ours outputs:}

\begin{lstlisting}[style=myprompt]
Updated Belief State:
- The contact "Sophie Martin" with number "+18723713947" has been successfully added and messaged.
- The contact "Olivia Alves" with number "+16278036185" has been successfully created.
- A message "hello, Olivia Alves" has been sent to her, and the current screen shows the SMS conversation, confirming the completion of the last iteration.
- All contacts in the list have now been processed, and the loop has finished.

Action:
WorkflowProgramCounterOperation.BREAK
\end{lstlisting}

\rev{\textbf{What is retained and what is pruned.} After the first iteration completes, \ours does \emph{not} keep every low-level action or screen transition. Instead, the retained state for the next iteration consists mainly of:
}\begin{enumerate}
    \item \rev{the STP;}
    \item \rev{the current PC;}
    \item \rev{a pruned low-level code history (completed loop iterations and untaken branches are folded);}
    \item \rev{the variables;}
    \item \rev{the global belief state.}
\end{enumerate}

\section{Prompts}
\label{sec:prompts}

\subsection{STP Generation Prompts}

\rev{Main Prompt of STP Generation:}
\begin{lstlisting}[style=myprompt]
{workflow_prompt}

# Your Job

You are a professional assistant trained to use the Task Program to complete diverse and complex tasks. The tasks can be related to data collection/processing, browser/device use, daily routine automation, etc.
You should carefully understand the user's requirement based on conversation, then write professional workflows in Task Program that can fulfill the user's needs. You are encouraged to do your job precisely and thoughtfully, beyond the user's expectations.

# Your Work Procedure

your goal is to generate a high-level workflow that summarizes the steps to complete the user-given task. The workflow should:

1. Follow the syntax of Task Program (a natural-language-style domain-specific language designed to describe workflows).
2. Decide whether and how to use the files, devices and apps in the workspace. Any non-existing file/device/app is not allowed in the workflow.
   Try to make the workflow generalizable (i.e. does not involve low-level operations that are only valid for specific data formats or UI design patterns).

# The Current Task

{task_description}

# Output Instruction


Note: 
1. Current user's language is English. When writing workflow/question/task_name, you need to use English to write.
2. **Do not** ask the user questions within the code. The user requires the task to be fully automated and will not participate in any part of the workflow. Therefore, asking or informing the user is unnecessary; you should complete the task autonomously.
3. Please set variable {answer} to as the final answer if the task requires you to answer a question.
4. Note: You must declare in which APP to operate. For example: `In "Contacts" App, add a new contact name 'Alice', and save it.` Otherwise, the workflow lacks accuracy, and the executor will not know which APP you need to use to complete the task.
5. The mobile phone is configured via a virtual image, and its date and time differ from the real-world date and time!! The date referred to in the Current Task is based on the **date on the mobile phone**, not the real-world date! Therefore, please make sure to emphasize using the mobile phone's date and time function to check today's date and time, instead of obtaining the real-world date via Python's datetime module! After obtaining "today's" date on the mobile phone, you can use the datetime module to calculate other dates based on the mobile phone's "today".

6. Remember to save! You must emphasize in the workflow that for tasks involving editing or creation, saving is mandatory at the end!

Now, Respond with the following content:

Your thought (starts with `--- Thought ---`), which contains your understanding about the task and plan of how to do the task in natural language.

The workflow (starts with `--- Workflow ---`), which is the workflow steps in Task Program syntax generated based on your understanding.

Example Format:
--- Thought --- (This contains your understanding about the task and plan of how to do the task in natural language.)
--- Workflow --- 
```
(Write the workflow steps in Task Program syntax generated based on your understanding.)
```

Please begin to write workflow!
\end{lstlisting}

\rev{Introduction prompt to STP, \ie, \textit{\{workflow\_prompt\}}:}

\begin{lstlisting}[style=myprompt]
I will introduce you to a new programming language called **Task Program**. This is a natural language style domain-specific language (DSL) used to describe task execution flows. You need to master this language and complete the tasks I arrange.

# Task Program Language Syntax

## 1. Basic Syntax and Concepts

The syntax of Task Program is similar to natural language, but following certain structures and concepts will help the system understand your intent more precisely.

A Task Program consists of a series of statements, usually one per line, describing an operation or a step, using imperative sentences or clear declarative sentences.

### Comments

Start a line with the `#` symbol to indicate that the line is a comment. Comments are used to explain code, and the system will ignore them during execution.

```
# This is a comment, the system will not execute this sentence
Tell user "Hello, comments make code easier to understand" # This is also an inline comment
```

### Variables

In Task Program, variables are represented by words wrapped in curly braces.

*   **Declaring (Recording) Variables**: Usually done using `record as {variable_name}` or similar natural language phrasing.

    ```
    Set variable {user_name} to "John"
    Set {current_city} to "Beijing"
    Create variable {visitor_count}, initial value is 0
    ```
*   **Using Variables**: Where stored information is needed, use `{variable_name}` to represent filling in the position with the variable's value.

    ```
    Tell user "Hello, {user_name}"
    Search in browser for "{current_city} population", record as {population_count}
    Tell user "The population of {current_city} is {population_count}."
    ```
*   Variable names should use meaningful words to facilitate understanding.

### Data Types

When variables are stored and passed in the system, they are understood as different types of data for actual operations. Different types of variables are used for data processing in different scenarios, such as number types for statistics, text types for document operations, dialogue, etc.

In Task Program, you do not need to specifically care about variable types; Task Program will automatically understand the type of each variable based on your code. Of course, you can also supplement the type description when defining variables to have more control over the entire process.

#### Basic Data Types

Task Program currently supports the following basic data types:

*   **Text**: A sequence of characters enclosed in quotes, such as `"Hello world!"` or `'Hello, World!'`.
*   **Number**: Integers or decimals, such as `10` or `3.14`.
*   **Boolean**: `True` or `False` (or `Yes` / `No`), used to represent conditions or states.

    ```
    Judge if {age} is greater than 18, record as {is_adult} # {is_adult} will be True or False
    If {is_adult} is true:
        Tell user "You are an adult."
    ```
*   **Image**: Represents image data. Can be obtained by reading files, etc.

    ```
    Read image file "logo.png", record as {company_logo}
    Tell user "This is the company Logo:"
    Send {company_logo} to user
    ```

#### Objects

An Object is a collection containing multiple **properties**, used to represent a specific thing or entity. Each property has a name and a corresponding value (also known as key-value pairs). The value of a property can be a basic type variable or another object.

*   **Creating Objects**:

    ```
    Create an object named {my_car}
    Set {color} of {my_car} to "Red"
    Set {brand} of {my_car} to "Tesla"
    Set {model} of {my_car} to "Model Y"

    # Or a more concise way, AI will understand your intent
    Create an object {client}, with attributes including: "name" is "Li Ming", "age" is 24, "city" is "Shanghai"
    ```
*   **Accessing Object Properties**:
    You can access property values directly using natural language inside curly braces, or use the dot `.` operator closer to traditional programming.

    ```
    Tell user "My car color is {color of my_car}"
    If {age of client} is greater than 18: # Assuming {client} is an object defined earlier
        Tell user "{name of client} is an adult."

    # Or using the {variable.property} format
    Tell user "My car color is: {my_car.color}, brand is {my_car.brand}, model is {my_car.model}"
    ```

#### Lists

A List is an ordered series of basic type elements or objects.

```
# Create list
Record list ["Apple", "Banana", "Orange"] as {fruit_list}
Count which public AI companies exist currently, output company name list, record as {company_list} # {company_list} is a text list

# Access list elements (AI will intelligently understand)
Get the first item of {fruit_list}, record as {fruit_1}
Tell user "My favorite fruit is: {fruit_1}"
Get the quantity of elements in {company_list}, record as {company_count}
Tell user "There are approximately {company_count} public AI companies."
```

#### Data Tables

Data Tables are similar to spreadsheets, containing rows and columns. **A Data Table can be viewed as a list composed of multiple objects with the same structure.** Each row in the table corresponds to an object, and each column corresponds to a property of the object. You can create data tables, add, delete, modify data in tables, import/export tables from/to files, etc.

```
# Create data table
Establish an empty data table containing columns "Name", "Department", "ID", record as {employee_info_table}

# Add a row (Can be understood as adding an object matching the structure)
Create an object {new_employee_A}, containing attributes "Name" is "Zhang San", "Department" is "Tech", "ID" is "T001"
Add {new_employee_A} as a new row to {employee_info_table}

# Add a row of content (More concise way)
Add a row of data {"Name": "Xiao Ming", "Department": "Marketing", "ID": "M023"} to {employee_info_table}

# Load data from file to table
Read content of table file "external_employees.xlsx", store into new data table {external_employee_table}

# Save data table to file
Save {employee_info_table} to file, filename is "employees.xlsx"
```

#### Data Processing

Task Program integrates AI large model capabilities. You only need to describe your processing requirements to complete various complex data processing operations.

*   **Filter**:
    ```
    Filter out all company records with "Market Cap" (Unit: Billion) greater than 100 from {company_data_table}, record result as {large_companies_table}
    ```
*   **Extract**:
    ```
    Get "Country" column from {company_data_table}, after deduplication record as {country_name_list}
    ```
*   **Summarize**:
    ```
    Calculate the sum of "Market Cap" for all companies in {company_data_table} where "Country" attribute is "China", record as {china_companies_total_cap}
    ```
*   **Sort**:
    ```
    Sort {company_data_table} by "Market Cap" column values from largest to smallest
    ```
*   **Merge**:
    ```
    Merge {domestic_sales_table} and {overseas_sales_table} by "Product ID" column, record as {global_sales_total_table}
    ```
*   **Calculate**:
    ```
    Divide "Market Cap" by "Employee Count" in {company_data_table}, add as "Per Capita Market Cap" column
    ```

### Flow Control

Code is executed sequentially line by line by default, but you can change the execution flow via flow control syntax.

#### Loops

When the same operation needs to be repeated for each item in a group, use a loop.

A loop consists of a loop condition and loop content. The loop condition describes the method and number of times, and the loop content is the steps to be repeated in each round.

The code for loop content usually needs indentation (leaving some spaces relative to the loop condition at the start of each line, usually 4 spaces) to indicate these operations are part of the loop. The Task Program AI interpreter has some tolerance for indentation, but good indentation helps readability and accuracy.

```
# Iterate through every {single_item_variable} in {data_collection}
Record list {"Apple Pie", "Banana Boat", "Orange Juice"} as {dessert_menu}
Iterate through every dessert in {dessert_menu}, record as {current_dessert}
    Tell user "We are selling: {current_dessert}"
```

#### Conditional Branching

When the program needs to perform different operations based on different situations, use conditional branching.

In conditional branching, the program decides which branch to execute based on the judgment of a condition expression.

A condition expression is generally a judgment, such as `{market_cap} less than 10 billion`, `{hospital_list} is empty`, etc., or a variable containing a boolean value.

*   **Basic Format**:

    ```
    If {condition_1}:
        ... (Execute when condition 1 is met)
    Otherwise if {condition_2}:    # Optional
        ... (Execute when condition 1 is not met but condition 2 is met)
    Otherwise:              # Optional
        ... (Execute when neither condition 1 nor 2 is met)
    ```

*   **Examples:**

    ```
    # Example 1
    If {user_age} is less than 18:
        Tell user "You are a minor."
    Otherwise if {user_age} is greater than or equal to 18 and {user_age} is less than 60:
        Tell user "You are an adult."
    Otherwise:
        Tell user "You have entered old age."

    # Example 2
    Query "How is the weather today?", output weather description, record as {weather_situation}
    If {weather_situation} contains "Rain":
        Tell user "Please remember to bring an umbrella when going out."
    Otherwise:
        Tell user "The weather is nice today!"
    ```

### Task Decomposition (Functions)

In Task Program, you can define a set of commonly used operations as a "Task", which is similar to "Functions" or "Subroutines" in traditional programming languages. This makes your main flow more concise and facilitates code reuse.

*   **Defining Tasks**:
    Start with `Define a task named "{task_name}"`. Tasks can have input parameters and outputs (return values).

    ```
    # Define a task to greet a specific user
    Define a task named "Custom Greeting"
        Task input: {user_name} (Type: Text)
        Record "Hello, {user_name}! Welcome to Task Program." as {greeting_text}
        Tell user {greeting_text}
        # This task has no explicit return value

    # Define a task to calculate sum of two numbers and return result
    Define a task named "Calculate Sum of Two Numbers"
        Task input: {number_one} (Number type) and {number_two} (Number type)
        Record result of {number_one} plus {number_two} as {calculation_result}
        Task returns {calculation_result} # Task ends, and outputs {calculation_result}
    ```

    The first line of the task definition or the lines immediately following are usually used to describe the required input parameters and their types (optional, but recommended).
*   **Calling Tasks**:
    Use `Execute task "{task_name}"`, provide necessary input parameters, and (optionally) receive output results.

    ```
    # Call greeting task
    Execute task "Custom Greeting", {user_name} is "Zhang San"

    # Call calculation task and get output
    Execute task "Calculate Sum of Two Numbers", {number_one} is 10, {number_two} is 25, output result record as {total}
    Tell user "Result of 10 + 25 is: {total}"
    ```

    The Task Program interpreter will intelligently match input parameter names. If parameter names in the definition and the call are not exactly the same but have similar meanings, the interpreter will also attempt to understand.

Using functions (tasks), you can decompose a complex large task into a series of smaller, easier-to-manage and describe sub-tasks, and then complete them one by one.
This ability to decompose tasks is essentially the ability of architecture and logic, requiring careful design, using variables to pass information at different stages of the flow, while ensuring the overall structure of the program is concise and readable.

## 2. Using Tools

Many complex functions rely on powerful tools to implement. The examples introduced above have actually involved some. This chapter will introduce them in detail.

### Operating Files

Task Program supports reading and saving files, including images, documents, tables, etc. Filenames are recommended to be wrapped in quotes; filenames are relative to the file path in the current IDE workspace.

```
Save text content "This is the first line written.\\nThis is the second line." to document "My Notes"
Read all text content of document "My Notes", record as {note_content}
Tell user "Note content is:\\n{note_content}"

Save {company_market_cap_table} as table file "company_market_cap_report"
Read data in table file "company_market_cap_report", record as {market_cap_table}

Check if file "thesis_draft" exists
If file exists:
    Read content of file "thesis_draft", record as {thesis_draft}
Otherwise:
    Tell user "Document does not exist."
```

Task Program will intelligently handle file read/write formats based on file suffixes and your description.

## 3. Task Program Writing Principles

1.  **Readability**: Program statements should be concise and smooth, easy for ordinary users to understand and modify.
2.  **Feasibility**: The program should fit the user's device and software environment, ensuring it can be executed correctly.
3.  **Generality**: Do not make any assumptions about things you don't know. Assumptions will definitely miss important things. You should use the generality of Task Program's natural language to generalize all possible situations. General language is still understandable to Task Program. For example, "Calculate xxx" is better than "If operation is addition, calculate a+b; if operation is subtraction, calculate a-b", because the former is more general.

## 4. Practical Cases

The following are representative examples showcasing the capabilities of the Task Program language in different scenarios.

### Case 1: Calculate average scores and grades for students based on their subject scores in the students.csv table (90 and above for A, 80-89 for B, 70-79 for C, 60-69 for D, below 60 for F), and save the results as student_grades.csv

```
# Read student grade table
Read the content of table file "students.csv" and store it in a data table, recorded as {student_grades_table}

# Analyze subject names (automatically identify score-related columns)
Analyze the subject names in {student_grades_table}, recorded as {subject_list}

# Add new columns for results
Add a new column "average_score" to {student_grades_table}
Add a new column "grade" to {student_grades_table}

# Iterate through each row to calculate average score and grade
Iterate through each row in {student_grades_table}, recorded as {current_student}:
    # Collect all scores for current student
    Create an empty list, recorded as {current_student_scores_list}
    Iterate through each subject name in {subject_list}, recorded as {subject_name}:
        Get the value of {subject_name} column from {current_student}, recorded as {current_score}
        Add {current_score} to {current_student_scores_list}

    # Calculate average score
    Calculate the average of all values in {current_student_scores_list}, recorded as {average_score}

    # Determine grade based on average score
    Calculate {grade} from {average_score}, 90 and above for A, 80-89 for B, 70-79 for C, 60-69 for D, below 60 for F

    # Store calculation results in current row
    Set the value of "average_score" column in {current_student} to {average_score}
    Set the value of "grade" column in {current_student} to {grade}

# Save results to new file
Save {student_grades_table} as table file "student_grades.csv"

# Display processing result summary
Get the number of rows in {student_grades_table}, recorded as {total_students}
Get the length of {subject_list}, recorded as {total_subjects}
Print "Grade processing completed! Processed {total_subjects} subject scores for {total_students} students, results saved as student_grades.csv"
```

### Case 2: Consolidating Contacts from table1.csv, table2.csv, and table3.csv into One Table, save as "contact_summary.csv"

```
# Read three contact table files
Read the content of table file "table1.csv" and store it in a data table, recorded as {contacts_table1}
Read the content of table file "table2.csv" and store it in a data table, recorded as {contacts_table2}
Read the content of table file "table3.csv" and store it in a data table, recorded as {contacts_table3}

# Create empty summary table
Create an empty data table, recorded as {contacts_summary_table}

# Get column structure from first table as base structure for summary table
Get all column names from {contacts_table1}, recorded as {column_names_list}
Iterate through each column name in {column_names_list}, recorded as {current_column_name}:
    Add new column {current_column_name} to {contacts_summary_table}

# Add data from table1 to summary table
Iterate through each row in {contacts_table1}, recorded as {current_contact}:
    Add {current_contact} as a new row to {contacts_summary_table}

# Add data from table2 to summary table
Iterate through each row in {contacts_table2}, recorded as {current_contact}:
    Add {current_contact} as a new row to {contacts_summary_table}

# Add data from table3 to summary table
Iterate through each row in {contacts_table3}, recorded as {current_contact}:
    Add {current_contact} as a new row to {contacts_summary_table}

# Remove duplicate contact records (based on name and phone number)
Remove duplicate rows from {contacts_summary_table} based on "name" and "phone" columns, keeping the first occurrence

# Sort by name
Sort {contacts_summary_table} by "name" column values from A to Z

# Save summary results
Save {contacts_summary_table} as table file "contacts_summary.csv"

# Count summary information
Get the number of rows in {contacts_summary_table}, recorded as {total_contacts}
Print "Contact consolidation completed! Organized {total_contacts} contacts in total, saved as contacts_summary.csv"
```

### Case 3: Calculate First N Terms of Fibonacci Sequence and Save to File

```
# Create data table to store results
Create an empty data table with "sequence_number" and "fibonacci_value" columns, recorded as {fibonacci_table}

# Initialize first two terms
If {terms_N} is greater than or equal to 1:
    Add a row of data {"sequence_number": 1, "fibonacci_value": 0} to {fibonacci_table}
    Record 0 as {previous_previous_term}

If {terms_N} is greater than or equal to 2:
    Add a row of data {"sequence_number": 2, "fibonacci_value": 1} to {fibonacci_table}
    Record 1 as {previous_term}

# Calculate terms 3 to N
If {terms_N} is greater than 2:
    Record 3 as {current_sequence_number}
    Repeat until {current_sequence_number} is greater than {terms_N}:
        # Calculate current term value
        Calculate the result of {previous_previous_term} + {previous_term}, recorded as {current_term}

        # Add result to table
        Add a row of data {"sequence_number": {current_sequence_number}, "fibonacci_value": {current_term}} to {fibonacci_table}

        # Update previous two terms for next calculation
        Record {previous_term} as {previous_previous_term}
        Record {current_term} as {previous_term}

        # Increment sequence number
        Calculate the result of {current_sequence_number} + 1, recorded as {current_sequence_number}

# Save results to file
Save {fibonacci_table} as table file "fibonacci.csv"

# Display calculation result summary
Get the "fibonacci_value" from the last row of {fibonacci_table}, recorded as {last_term_value}
Tell user "Calculation completed! The {terms_N}th term of Fibonacci sequence is: {last_term_value}"
Tell user "Complete results have been saved to fibonacci.csv file"

```

### Case 4: Batch Processing Student Grade Data and Generating Individual Grade Reports, subjects include "Math", "Chinese", "English"

```
# Define task: Calculate individual student grade statistics
Define a task named "calculate_student_grade_statistics"
    Task input: {math_score} (number type), {chinese_score} (number type), {english_score} (number type)

    # Calculate total score
    Calculate the result of {math_score} + {chinese_score} + {english_score}, recorded as {total_score}

    # Calculate average score
    Calculate the result of {total_score} / 3, recorded as {average_score}

    # Generate comment based on average score
    If {average_score} is greater than or equal to 90:
        Record "Excellent! Keep it up!" as {comment}
    Otherwise if {average_score} is greater than or equal to 80:
        Record "Good, there's room for improvement." as {comment}
    Otherwise if {average_score} is greater than or equal to 70:
        Record "Average, need to strengthen learning." as {comment}
    Otherwise if {average_score} is greater than or equal to 60:
        Record "Pass, but need to work harder." as {comment}
    Otherwise:
        Record "Fail, need to focus on learning." as {comment}

    # Create grade statistics object
    Create an object {grade_statistics} with attributes: "total_score" is {total_score}, "average_score" is {average_score}, "comment" is {comment}

    Task returns {grade_statistics}

# Define task: Generate individual grade report text
Define a task named "generate_grade_report"
    Task input: {student_name} (text type), {grade_statistics_object} (object type), {class_rank} (number type)

    # Extract grade statistics information
    Get the "total_score" attribute from {grade_statistics_object}, recorded as {total_score}
    Get the "average_score" attribute from {grade_statistics_object}, recorded as {average_score}
    Get the "comment" attribute from {grade_statistics_object}, recorded as {comment}

    # Generate report text
    Combine the following content into report text, recorded as {report_content}: "=== {student_name} Individual Grade Report ===\\n", "Total Score: {total_score} points\\n", "Average Score: {average_score} points\\n", "Class Rank: #{class_rank}\\n", "Comment: {comment}\\n", "Report Generation Time: current date time\\n", "================================\\n" # Note, do not line break

    Task returns {report_content}

# Main program: Batch process student grades
# Read student grade table
Read the content of table file "student_scores.csv" and store it in a data table, recorded as {student_grades_table}

# Create result table to store all students' statistics
Create an empty data table with "name", "total_score", "average_score", "rank", "comment" columns, recorded as {grade_statistics_table}

# First round: Calculate grade statistics for each student
Iterate through each row in {student_grades_table}, recorded as {current_student}:
    # Get student information
    Get the value of "name" column from {current_student}, recorded as {student_name}
    Get the value of "math" column from {current_student}, recorded as {math_score}
    Get the value of "chinese" column from {current_student}, recorded as {chinese_score}
    Get the value of "english" column from {current_student}, recorded as {english_score}

    # Call task to calculate grade statistics
    Execute task "calculate_student_grade_statistics", {math_score} is {math_score}, {chinese_score} is {chinese_score}, {english_score} is {english_score}, output result recorded as {grade_statistics} # Function call do not line break

    # Add statistics results to statistics table
    Add a row of data to {grade_statistics_table}, "name" is {student_name}, "total_score" is {grade_statistics.total_score}, "average_score" is {grade_statistics.average_score}, "comment" is {grade_statistics.comment}

# Sort by total score and add ranking
Sort {grade_statistics_table} by "total_score" column values from high to low
Record 1 as {current_rank}
Iterate through each row in {grade_statistics_table}, recorded as {current_row}:
    Set the value of "rank" column in {current_row} to {current_rank}
    Calculate the result of {current_rank} + 1, recorded as {current_rank}

# Second round: Generate individual report files for each student
Iterate through each row in {grade_statistics_table}, recorded as {current_statistics}:
    # Get student information
    Get the value of "name" column from {current_statistics}, recorded as {student_name}
    Get the value of "rank" column from {current_statistics}, recorded as {class_rank}

    # Rebuild grade statistics object (for passing to report generation task)
    Create an object {current_grade_statistics} with attributes: "total_score" is the value of "total_score" column from {current_statistics}, "average_score" is the value of "average_score" column from {current_statistics}, "comment" is the value of "comment" column from {current_statistics}

    # Call task to generate report
    Execute task "generate_grade_report", {student_name} is {student_name}, {grade_statistics_object} is {current_grade_statistics}, {class_rank} is {class_rank}, output result recorded as {individual_report} # Note, one line of code cannot be written across lines

    # Save individual report to file
    Save text content {individual_report} to document "{student_name}_grade_report.txt"

# Save summary statistics table
Save {grade_statistics_table} as table file "class_summary.csv"

# Completion notification
Get the number of rows in {grade_statistics_table}, recorded as {total_students}
Tell user "Batch processing completed! Generated individual grade reports for {total_students} students, summary table saved as class_summary.csv"
```

### Case 5: Minimum Spanning Tree

```
Define a task named "compute_minimum_spanning_tree"
    Task input: {n} (number type), {edges} (list type, each element is a tuple of three positive numbers representing the start node, end node, and edge weight)

    Solve the following problem: "Given a weighted undirected graph with n nodes, construct a Minimum Spanning Tree (MST) that connects all the nodes with the minimal total edge weight. Each edge can be used only once, and no cycles are allowed. If the graph is disconnected and an MST cannot be formed, return -1.". Record the calculated total weight of the minimum spanning tree as {mst_total_weight} # Note, the task description must be written on one line.

    Task returns {mst_total_weight}
```

### Case 6: Add the following tasks into the Todoist app: "Task: Complete quarterly report  due_date: 2025-09-15  priority: High  project: Work  note: Include sales data and projections", "Task: Schedule dentist appointment  due_date: 2025-09-20  priority: Medium  project: Personal  note: Call Dr. Wilson's office", "Task: Buy groceries for dinner party  due_date: 2025-09-13  priority: High  project: Home  note: Need ingredients for pasta and salad".

```
Record "Task: Complete quarterly report  due_date: 2025-09-15  priority: High  project: Work  note: Include sales data and projections", "Task: Schedule dentist appointment  due_date: 2025-09-20  priority: Medium  project: Personal  note: Call Dr. Wilson's office", "Task: Buy groceries for dinner party  due_date: 2025-09-13  priority: High  project: Home  note: Need ingredients for pasta and salad" as the {task_list}.
Iterate over the {task_list}, and for each {task_item}, do the following:
    In the `Todoist` app, Navigate to the task creation page, fill the form with the {task_item} information including title, due date, priority level, project category, and notes, then save the task. # Note, here you must write "In the `Todoist` app" to ensure workflow accuracy. Also, remember to save.
```

### Case 7: Delete All Contacts whose name starts with 'A' in the `Contacts` app.

```
In the `Contacts` app, Delete All Contacts whose name starts with 'A'. # CRUD operations are universal atomic operations, no need to write them too complicatedly! The more complex, the easier to error! Deleting items doesn't need a loop, just one sentence, the executor will automatically complete this task.
```

## 5. Task Program Writing Suggestions

In long-term and extensive Task Program programming practice, we have summarized the following programming suggestions. Following these suggestions can make your Task Program code more stable, reliable, versatile, and powerful.

1.  **Avoid fabrication and assumptions (Very Important!!)**. Sometimes the information provided by the user is not comprehensive, for example, table analysis without giving the specific content of the table. In this case, **you must use existing information as much as possible, instead of fantasizing without basis that a certain field exists in the table**, or that a certain column of the table must have specific content. These fabrications and assumptions do not work in real problems. **Since Task Program language is dynamic, you can let Task Program autonomously analyze what exists in the table**. Just as shown in our cases (Analyze the subject names in {student_grades_table}, recorded as {subject_list}, because we don't know which subjects are in the table, we can let Task Program automatically analyze which subjects are there, instead of presumptuously assuming "Chinese", "Math", etc. exist. In this example, the input of "Analysis" is a table, and the output is a list of subjects. It is visible that we can let Task Program autonomously analyze the table to obtain required information.)
2.  **Avoid trivial details, focus on core goals of the task**, making the code universal. For example, "Go to the convenience store" is better than "Walk 40 meters east, then 20 meters north", because the goal is to buy things, just caring about whether the store is reached is enough, no need to give detailed navigation operations. These low-level operations should be left for execution time, not programming time. The program should focus more on the work logic itself, trying to discard things unrelated to the core flow.
3.  **Consider conciseness**. For example, if there is no conflict in the table, try to traverse it in one loop instead of writing multiple redundant loops. Of course, if multiple loops are necessary, they should be written appropriately.
4.  **One operation per line**. Task Program syntax requires us that **the same operation cannot be written across lines**, strings, attributes, etc., must be written on the same line, do not line break.
5.  **When encountering algorithm problems**, just define a corresponding task to solve it, describe the task framework roughly, do not define specific implementation. Algorithm tasks only need to define input, task description, return value, and the Task Program interpreter will automatically implement relevant algorithms.
6.  **If it involves defining classes, objects, data structures**, lines must be written at the very beginning. First write class/object definitions, then task definitions, finally the workflow.
7.  **Must follow the example format to write Task Program!** Do not add labels like 1.2.3.4 before the Workflow!
8.  **Note, statements requiring APP operation must declare operation in xxx APP**, for example: `In "Contacts" App, add a new contact name 'Alice'.` Otherwise, the workflow lacks accuracy, and the executor doesn't know which APP you need to use to complete the task.
9.  **Create, Read, Update, Delete (CRUD) are all universal atomic operations**, no need to write out their underlying specific behaviors, just one sentence is enough! The more complex, the easier to error! Deleting items does not need to be written as an underlying loop, just one sentence, the executor will automatically complete this task, specifically how to delete is left to the executor's underlying layer to complete!
\end{lstlisting}

\subsection{STP Execution Prompts}

\begin{lstlisting}[style=myprompt]
# Introduction

You are an expert in solving problems with python. Your job is to complete given tasks by tranlating pseudo code (natural language-style step-by-step workflow description) to python code lines that can be executed reliably on the fly.

You will be given:

- **(1) The task description**, such as "read the math expressions in each line of data.txt", "read the emails and add relevant events to calendar", etc.
- **(2) The current workflow context**, including the pesudo code of the workflow, the current position of workflow code (marked as "current step" in inline comment).
- **(3) The current python context**, including the recent python code lines generated and executed.
- **(4) The available variables and files** in the current context and their current values (or brief summary of values).

You will be requested to run in two mode:

**Mode 1: Code Generation**

In this mode, you should output the **next lines of python code** to complete the current workflow step (marked as "current step" in inline comment) according to the task and context. The new python code lines should follow the principles below:

- **(1) Correctness**: The code will be executed immediately after being generated, so it must be a valid line of python code (do not use undefined variables; do not perform infeasible operations on the current data).
- **(2) Simplicity**: You are encouraged to break the workflow step into multiple python steps; Keep each python step simple and atomic. You should only output the python code lines to complete the current workflow step, no other content.
- **(3) Data awareness**: You are provided with the data schema and values, so your code should be closely customized to the data, you can directly use the data values (or any values inferred from the data) in the python code. You don't have to worry about whether the generated code can be generalized or reused.
- **(4) Sequential execution**: The control-flow structures (loops, if-else, etc.) in the workflow should be interpreted as sequential steps in python, i.e. you decide which branch to take based on the loop/if-else conditions.

**Mode 2: Workflow Status Update**

In this mode, you should decide how to update the workflow program counter (WPC) based on the execution output of your generated python code. The WPC operations include:

- **hold**: stay in the current workflow step if it has not been completed.
- **continue**: move to the next workflow step.
- **break**: jump out of the current loop in the workflow.
- **return**: jump out of the current function in the workflow.

** Important tips

- One Step at a Time: Only write the Python code for the single task marked with a comment like # <-- current step. Do not try to solve future steps.
- No Complex Logic: Do not write full Python loops (for ...:) or conditionals (if ...:) yourself. The plan already contains the logic. You will be guided through loops and conditionals one piece at a time.
- When you encounter the loop for the first time, your job is only to set up the loop and get the very first item. When you return to the loop step later, your job is to increment the counter and get the next item.
----

{example_prompt}

# Code Generation Guidelines

Through long-term and extensive code generation practice, we have summarized the following programming suggestions. Following these suggestions will enable you to write code that is more stable, reliable, versatile, and powerful.

1. Focus on the step of the current line. Focus on iteration for loop steps, judgment for conditional steps, and the action of the current step for sequential execution steps. Do not overdo it; do not perform steps that are required later in advance, and do not miss or underperform steps. Do not casually use variables with unknown context sources; do exactly what needs to be done for the current step. Do not be lazy; use the safest solution to ensure the current step is completed.
2. **Make good use of Large Language Model calls, prioritizing the use of the LLM (llm.query) to analyze data and enhance program robustness**. Sometimes you need to handle irregular data or uncertain situations, analyze them, or unify their formats. Therefore, we provide you with the LLM API call `llm.query` to handle such irregular data. If processing data programmatically always results in errors, you can directly ask the LLM to help you integrate information and extract formats! The Large Language Model is the safest solution!
3. Use dictionaries and lists to define data structures. **Do not define custom Classes!** Dictionaries and lists are sufficiently complete to describe a structure. For example, a person can be defined as: `{"name": "Alice", "age": 12}`.
4. If an error occurs during programmatic processing, please try **using llm.query to integrate information**!! Note that `llm.query` can only return JSON-like objects, so it only supports basic types or combinations of dictionaries and lists (str, int, bool, float, list, dict; any other types are not supported, and `object` cannot be used; `{str: object}` is an incorrect syntax!). You cannot request it to return other types of Python objects.
5. **`if` statements bring execution instability, so do not use them**. Except for specially permitted cases, do not use `if` statements; use `judge = ...` instead! For example, the following usage is wrong: `if <judge_expression>:`. The following usage is correct: `judge = <judge_expression>`. **Do not use the `if judge: ...` statement either; you simply cannot have any `if`!** You judge whether to execute branch operations by directly reading the value of `judge` (True, False) after calculating it, rather than using `if`.
6. **Carefully check if `current_screenshot` meets the requirements!** When executing the workflow, relate it to the context and think about the purpose of the current step; do not look only at the current step! Sometimes the code seems complete, but the screenshot clearly shows the goal hasn't been reached; you need to be strict with yourself! If there are potential flaws or unreasonable aspects in the interface (e.g., inputting text in an inappropriate place, clicking submit/send buttons with no actual effect), please strictly output `hold`!
7. When executing the workflow, relate it to the context and think about the purpose of the current step; do not look only at the current step!
8. Note that everything is based on the environment (`current_screenshot`) as the factual baseline! The Belief State might be outdated, and executed Python code might contain implicit execution errors. You can only faithfully observe everything based on the environment; seeing is believing. If there is indeed a problem in the `current_screenshot`, it means the previous Python code execution failed or encountered other unexpected changes. Do not be misled by the previous Python code!
9. Strictly follow the Belief State Plan requirements. If the content required by the Plan and the Current Line differs, prioritize the Plan's requirements.

# Framework Usage

To complete the task, you should make the most use of the APIs included in the framework we provide to you. Here is a document to teach you how to use this framework. If no framework provided, you can use the standard python function only.

{framework_prompt}

Note: Do not call non-existent APIs that are not described in the document above.

----


# Supplementary Instructions on Belief State

Based on the files and case requirements above, you also need to generate content related to the Belief State.

In the task, you are facing a partially observable environment (such as a mobile phone, computer, or browser interface). In such an environment, you can only observe part of the environmental status, while other parts remain hidden. Therefore, you need to attempt to construct and maintain a Belief State to infer and explain the complete mechanism of the environment and the hidden variables within it, constantly updating and correcting the Belief State through environmental interaction to facilitate further planning to complete the task.

First, you need to summarize the Belief State based on the task background, goals, environmental observations, and interaction history. The Belief State includes analysis and summary of the current environmental state, analysis of environmental mechanisms, and speculation on possible future states of the environment under certain meaningful operations. In the Belief State, you need to explain the approximate interaction logic of the environment based on task requirements, environmental state, and interaction history, combined with your experiential knowledge. This content needs to be relevant to achieving the task goals and can include confirmed theories (conforming to general theories or confirmed by interaction history) and unconfirmed hypotheses. However, you must clearly indicate which are confirmed or conform to general laws and which are merely hypotheses. Hypotheses can contain multiple possibilities.

Structurally, the Belief State is a list, where each entry is a faithful record or hypothesis of the environmental state. For example:
- The `name` field of the contact is set to "John Smith".
- There is a menu icon on the top right of the contact page, opening this menu *might* lead to the `delete contact` interface.

Each item needs to be specific, detailed, and atomic, starting from what you observe. It can include both deterministic conclusions and your speculations and hypotheses. You must inherit and modify the previously existing Belief State. **Do not arbitrarily modify the previous Belief State!** They are the best summary and analysis of the hidden state of the environment. Arbitrarily deleting them will cause you to lose understanding of the environment, so please retain every original text as faithfully as possible, correcting or discarding it only when it clearly contradicts the real environment.

Note:
- For information in the Belief State that contradicts the `current_screenshot`, the Belief State needs to be corrected to be consistent with the `current_screenshot`;
- For Belief State entries that cannot be directly judged as true or false from the `current_screenshot`, do not modify the Belief State for now. When you update the Belief State, be sure to keep these entries! For example, Belief State before update: [a, b, c]. If observation reveals that entry `a` does not match reality, entry `b` matches reality, and entry `c` cannot be determined, then the updated Belief State: [b, c]. `a` does not match the environmental situation and needs to be removed or corrected; `b` needs to be retained as it is still correct; `c` *must be retained*. Although it is not reflected in the current `current_screenshot`, it implies it might be hidden later. This `c` is the most critical information because you cannot read it directly from the screen. If you omit it, you cannot find it again, which will affect the progress of your subsequent tasks.

Next, you need to create a plan:
1. If there are still unconfirmed hypotheses in the current Belief State that are crucial for completing the task, you can create a plan to further explore the environment to verify or eliminate the hypotheses;
2. If the current Belief State is sufficient to complete the current task goal, you can attempt to create a plan to solve the current task (only addressing the task of the current line, or handling unexpected situations when necessary);
3. If you observe during the interaction that the current task has been completed, you can indicate in the plan that the task is finished.

Note:
1. As the interaction proceeds, the environmental state updates constantly; you must keep the Belief State up to date.
2. Everything is based on the environment. The Belief State might be outdated, and executed Python code might contain implicit execution errors. You can only faithfully observe everything based on the environment. You must constantly check if the existing Belief State matches the real situation in the environment. Sometimes the current Belief State contradicts the real environment and interaction history. This indicates the previous Belief State was incorrect; you should try to overturn it and rewrite the Belief State and corresponding plan starting from the real environment;
3. Sometimes the environment is subject to **unknown external interference** or certain execution errors, causing the environment to deviate from the state analyzed in the Belief State! For example, previous operations were disrupted by certain factors! At this point, interaction history may not truly reflect the real situation of the environment; execution errors might have occurred in the interaction code. You need to keenly detect these changes and be sure to reasonably correct the analysis and planning content of the Belief State. For instance, if text in a form is different from what was in the previous Belief State, or a file that existed originally is now missing, you must pay attention to these unexpected changes.
4. If unexpected changes occur that sabotage previous operations, you need to **immediately** pause the current task, fix the unexpected situation, and then come back to continue the current task, instead of saying "fix it later." You must **pause the execution of the current workflow step and switch to attempting to fix the unexpected/abnormal situation**.
5. You can choose an appropriate exploration-exploitation trade-off strategy, verifying hypotheses while utilizing the results in the Belief State to complete the task;
6. If no environmental interaction is involved (e.g., code operations, setting variables, etc.), you do not need to analyze the Belief State.
7. Your plan only needs to contain the action for the next step; you do not need to plan multiple steps. It only requires a clear command action. This action must be atomic and a simple sentence, not a coordination of multiple actions! For example: "Open the 'Contacts' app" counts as one action and can be written this way; however, "Open the Contacts app and create a new contact" constitutes two actions and cannot be written this way.
8. Regardless of whether it is the current step, you should proactively observe environmental anomalies rather than rigidly following the workflow step by step. Since a problem in any step of the workflow will lead to task failure, please take responsibility and care about the global situation, not just the current step. Note: If you detect that the workflow execution is disrupted by any unexpected situation, please modify the Plan to pause the current workflow step execution, prioritize handling the unexpected situation to get the workflow back on track, and then resume execution.
9. The following statement is incorrect and must be absolutely avoided:

```
The incorrect step xxx should be addressed in its own step if required, but does not block progress for the current step.
```

Your responsibility is to monitor the correct execution of the entire workflow! It does not mean that if your current step is done, you don't need to care whether the previous steps were completed! Please bear this in mind. The correct attitude is:

```
I take full responsibility for monitoring the entire workflow to ensure it runs correctly from start to finish. 
Before proceeding with my current step, I always verify that all previous steps have been completed correctly. 
If I find any issue in earlier steps, I must stop and address it immediately before moving forward. 
My goal is not only to complete my own step, but to maintain the integrity and success of the whole workflow.
```

## Examples for Belief State 

### Example 1 (During Normal Execution)
#### Input
Task: Create a contact named agent prog in `Contacts` app and set email to '123@test.com'.
current line: Create a contact named agent prog in `Contacts` app.

current belief state:
- We are now at the home page of the mobile phone.
- `Contacts` app is exist on the phone.
- The contact app may support creating contacts through a “+” or “Create contact” button.
- Once inside the creation page, text fields like “Name” and “Email” might become interactable.
- There may be permission pop-ups (e.g., “Allow Contacts to access photos”) blocking the UI if I open the Contact Creation Page, which could explain missing buttons.

Mode: Code Generation

#### Output
--- Observation ---
The screenshot shows the main page of `Contacts` app, and there is a '+' floating button shown on the screen. There is no contact in the contact list. At the bottom part of the app, there are 4 tabs including "contacts", ...(Every element, especially the icons and texts, should be mentioned in your response. )

--- Thought ---
The task requires creating a new contact named "agent prog" in the Contacts app. The current observation indicates that the Contacts app is open. So `We are now at the home page of the mobile phone.` is wrong and should be discarded. ``Contacts` app is exist on the phone.` is True, so it should be kept ALWAYS. There is a '+' button so `The contact app may support creating contacts through a "+" or "Create contact" button.` should be revised; Although we don't know whether `Once inside the creation page, text fields like "Name" and "Email" might become interactable.` is correct, it should be kept until there is an opposite observation against this, so do the last entry (There may be permission pop-ups (e.g., "Allow Contacts to access photos") blocking the UI if I open the Contact Creation Page, which could explain missing buttons.). They should be kept.

--- Updated Belief State ---
- The contact app supports creating contacts through a "+" button.
- `Contacts` app is exist on the phone.
- Once inside the creation page, text fields like "Name" and "Email" might become interactable.
- There may be permission pop-ups (e.g., "Allow Contacts to access photos") blocking the UI if I open the Contact Creation Page, which could explain missing buttons.

--- Judgement ---
There is no issue shown at this time, so I should considering the current step "Create a contact named agent prog in `Contacts` app.". I will plan to enter the Contact Creation Page of Contacts app.

--- Plan ---
Click on the "+" button. (For Code generation Mode, You should output the description of next code action; For Workflow Status Update Mode, You should output what wpc operation you will do next.)

--- Action ---
```
mobile.click("'+' button")
```

## Example 2 (When Suspicious Issues Occur)

#### Input

Task: Create a contact named agent prog in `Contacts` app and set email to '123@test.com'.
current line: Create a contact named agent prog in `Contacts` app.

current belief state:

* We are now at the home page of the mobile phone.
* `Contacts` app is exist on the phone.
* The contact app may support creating contacts through a “+” or “Create contact” button.
* Once inside the creation page, text fields like “Name” and “Email” might become interactable.
* There may be permission pop-ups (e.g., “Allow Contacts to access photos”) blocking the UI if I open the Contact Creation Page, which could explain missing buttons.

Mode: Workflow Status Update.

#### Output

--- Observation ---
The screenshot shows a system pop-up that says "Contacts keeps stopping." with two buttons: "App info" and "Close app". The main UI of the Contacts app is not visible.

--- Thought ---
The observation indicates that the `Contacts` app has crashed and cannot proceed to the creation page. Therefore, `The contact app may support creating contacts through a "+" or "Create contact" button.` cannot be verified for now, and the current step cannot continue. `Contacts app is exist on the phone.` remains true, but the app is currently not in a functional state. The belief `We are now at the home page of the mobile phone.` is false because the phone is displaying a crash dialog instead. I must resolve this issue before proceeding.

--- Updated Belief State ---
* The screen shows a system pop-up "Contacts keeps stopping." blocking any further interaction.
* `Contacts` app is exist on the phone but is currently not responding (crashed).
* The contact app may support creating contacts through a "+" or "Create contact" button.
* Once inside the creation page, text fields like "Name" and "Email" might become interactable.
* There may be permission pop-ups (e.g., "Allow Contacts to access photos") blocking the UI if I open the Contact Creation Page, which could explain missing buttons.

--- Judgement ---
The workflow cannot continue because the current app is not functioning properly. I must not proceed with creating the contact until the issue is resolved.

--- Plan ---
Try to restart the Contacts App. Before that, I cannot proceed to next step.

--- Action ---
hold

# Output Format

--- Observation ---
(Faithfully, comprehensively, and exhaustively record what you see on the interface. Only say what you see; record only the facts you see. Do not record any of your value judgments. For example, terms like "the previous step was executed correctly," "meets expectations," "the previous step was executed incorrectly," etc., are value judgments and must not appear; describe fully what you see on the screen. Describe what you see on the current screen in as much detail as possible. Every element, especially the icons, texts, should be mentioned in your response.)

--- Thought ---
(Your thoughts, which contains your understanding about the observation, and how to update belief state, next plan and action. Pay attention to detecting all possible anomalies or unexpected situations from the Observation! If a situation occurs that does not meet workflow expectations, you need to update the Belief State. Even if an error occurred in a previous step, you need to correct it immediately in the current situation, **pause the execution of the current workflow step, switch to attempting to fix the unexpected/abnormal situation**, and attempt to handle these unexpected and abnormal situations in the Plan.)

--- Updated Belief State ---
(The new Belief State, in natural language, as required above. It needs to be corrected based on environmental facts while inheriting the previous Belief State, and all previous details should be preserved as much as possible.)

--- Judgement ---
(Output a judgment on the current global situation. If suspicious errors or unexpected errors are raised in the Belief State, you should pause the execution of the workflow plan, switch to dealing with these suspicious issues, and make the correct choice in the Plan.)

--- Plan ---
(Description about what you want to do next single step. Since a problem in any step of the workflow will lead to task failure, please take responsibility and care about the global situation, not just the current step.)

--- Action ---
(In Code Generation Mode, you should output code as action, which is the actual script in Python syntax generated based on your understanding and the task requirements; Or wpc operations including `continue`, `hold`, `break`, `return` in Workflow Status Update Mode. In Workflow Status Update Mode, you can only output one of `hold`, `continue`, `break`, `return` as the Action.)

# Your Turn!

Now, please generate the python code lines based on the following task and context.

Task: {task_description}

Workflow context:
```
{workflow_context}
```

Data & variables:
{data_and_variables}

Python context:
```
{python_context}
```
Note: Python Context is your operation history in the environment, but these executed operations may not have succeeded in the environment. There are likely implicit errors, or the environment may have changed due to external factors, rendering previous operations void. At this point, you need to **pause the execution of the current workflow step and switch to attempting to fix the unexpected/abnormal situation**. Therefore, **everything depends on the environment; everything depends on the content observed in current_screenshot; the operation history is for reference only.**

Mode: {generation_mode}

The current belief state: 
```
{belief_state}
```
In Code Generation And Workflow Status Update Mode, you should first observe the environment, update and correct the information in the Belief State based on the environment, and provide the next Plan accordingly. Note: If you detect that the workflow execution has been disrupted by any unexpected situation (regardless of whether it is at the current step; you should proactively observe environmental anomalies rather than rigidly following the workflow step by step. Since a problem in any step of the workflow will lead to task failure, please take responsibility and care about the global situation, not just the current step), please modify the Plan to pause the execution of the current workflow step, prioritize handling the unexpected situation to bring the workflow back on track, and then resume execution.

{additional_info}
'''

\end{lstlisting}

\subsection{Action Space}

Action Space, \ie, \textit{\{framkwork\_prompt\}}:

\begin{lstlisting}[style=myprompt]
- **mobile**: Primitives for manipulating mobile devices.
  - The following APIs are only available for Android devices:
    - `mobile.start_app(app_name: str)`: Open the app named `app_name`.
    - `mobile.kill_app(app_name: str)`: Kill the app named `app_name`.
  - The following APIs are available for all devices:
    - `mobile.get_input_field_text(view_description: str) -> str`: Get the text from the input field specified by `view_description`.
    - `mobile.get_clipboard()`: Get the text from the clipboard.
    - `mobile.set_clipboard(text: str)`: Set the text to the clipboard.
    - `mobile.expand_notification_panel()`: Expand the notification panel. This includes Internet, Bluetooth, Flashlight, and so on.
    - `mobile.take_screenshot() -> PIL.Image`: Take a screenshot of the current screen and return a PIL image. Note: This is not for "taking a photo" using the mobile phone. For taking photos, you should use the camera app instead.
    - `mobile.back()`: Navigate back from the current screen.
    - `mobile.home()`: Navigate to the home screen.
    - `mobile.back_to(description: str, max_steps=5)`: Navigate back to the view described by `description`. `max_steps` is the maximum number of steps to navigate.
    - `mobile.swipe_upward(view_description: str, distance=None)`: Swipe up on the view specified by `view_description`. `distance` is a number to control the distance for the swipe action, e.g., `mobile.swipe_upward(view_description="scroll_view", distance=100)`.
    - `mobile.swipe_downward(view_description: str, distance=None)`: Swipe down on the view specified by `view_description`. `distance` is a number to control the distance for the swipe action.
    - `mobile.swipe_leftward(view_description: str, distance=None)`: Swipe left on the view specified by `view_description`. `distance` is a number to control the distance for the swipe action.
    - `mobile.swipe_rightward(view_description: str, distance=None)`: Swipe right on the view specified by `view_description`. `distance` is a number to control the distance for the swipe action.
    - `mobile.swipe_until(view_description: str, expected_desc: str, towards: "up", duration=1000, max_retry: int = 10) -> bool`: Swipe the view specified by `view_description` until `expected_desc` is fulfilled. If the desired view appears, return True; otherwise, return False.
    - `mobile.wait_until(description: str, waitInterval:float=0.5, timeout=5) -> bool`: Wait for a view described by `description` to appear and return it. `timeout` is the time limit (in seconds) for waiting. `-1` means unlimited. If the desired view appears, return True; otherwise, return False.
    - `mobile.check(description: str) -> bool`: Check whether the current screen state matches `description`. If matched, returns True; otherwise, returns False.
    - `mobile.click(view_description: str)`: Click the view specified by `view_description`.
    - `mobile.input(view_description: str, text: str)`: Clear the input field specified by `view_description` and input the given text. You don't have to call a keyboard; use this input method directly.
    - `mobile.input_by_pasting(view_description: str, text: str)`: Input text into the view specified by `view_description` by pasting. Use the `input_by_pasting` API when standard input doesn't work, such as in the WeChat app.
    - `mobile.long_click(view_description: str)`: Long click the view specified by `view_description` for 1 second.
    - **Note**: UI operations are distinct. In Code Generation mode, it is best to generate only one step at a time for these UI operations because every UI operation changes the environment. You should proceed step-by-step; otherwise, errors can easily occur.

- **llm**: The interfaces to get answers from foundation models.
    - `llm.query(prompt_or_image1, ..., prompt_or_imagen, returns)`: Query a large language model. Positional arguments can be a list of string prompts or images. The keyword parameter `returns` specifies the expected return type and description for function outputs. It supports the following formats:
        1.  **Single Typed Return**:
            - Returns a single value parsed according to the specified type.
            - First element: Description of the return value.
            - Second element: Expected type of the return value.
            - Example: `returns=("age", int)` -> function returns the parsed age value as an integer.
            - Example: `returns=("items", list[str])` -> function returns a list of strings representing the items.

        2.  **Multiple Returns**: (Pay close attention: the outer layer must be wrapped in a list, while the description is wrapped in a tuple!!)
            - Returns multiple values with optional type specifications.
            - Each element can be either:
                a) str: Description only (assumes str type).
                b) tuple: (description, type) pair.
            - Example: `returns=["name", ("age", int)]` # For instance, here the return value is defined as two elements. The first is a string described as "name" (equivalent to `("name", str)`). The second is an integer described as "age" (note it is `("age", int)`, not `["age", int]`!!). If your `llm.query` call fails, you must reflect on whether you met the format requirements I stated.
                -> `[string_name, parsed_integer_age]`
            - Example: `returns=[("items", list[str]), ("count", int)]`
                -> `[parsed_list_of_strings, parsed_integer]`
            - Specific usage example: `reign_name, reign_period, main_events, contemporary_west, cultural_exchange = llm.query(f"Please analyze the raw info into desired fields:", str(raw_history_info), returns=[("era name", str), ("reign period", str), ("major events", str), ("contemporary Western history", str), ("Sino-Western cultural exchange events", str)])`. **Note**: You must input the data information you want to analyze as a string or image; do not just give a prompt without data. Without data (like `raw_history_info` in the example), the large model cannot answer correctly.

        3.  **Complex Requests**: Nested lists, dictionaries, and `llm.query` support these as well. Reference documentation follows:
            - `llm` supports all the following request types:
            3.1. **Basic Types**:
            Supports `bool`, `int`, `float`, `str`, constants (e.g., `"test"`).

            3.2. **Sum of Basic Types**:
            Supports sum type declarations like `bool | int`.

            3.3. **Homogeneous List Type**:
            e.g., `list[str]` indicates a list where the element type is `str`.

            3.4. **Homogeneous Key, Homogeneous Value Dictionary Type**:
            e.g., `dict[str, int]` indicates a dictionary with keys of type `str` and values of type `int`, with arbitrary length.

            3.5. **Nested Lists**:
            If you need finer control over lists, you can replace `list` with `[<type spec>]`. The nested element type `<type spec>` can be any type conforming to the rules in this document.

            3.5.1 **Fixed Length List**:
            e.g., `[str, int, int]` indicates a list of length 3, where the first element is `str`, the second is `int`, and the third is `int`.

            3.5.2 **Variable Length List**:
            The list length is not fixed. In this case, the last element of the list must be `...`. e.g., `[int, str, ...]` indicates a list of arbitrary length where each element can be `int` or `str`.
            Using `[int | str, ...]` also conveys this meaning.

            3.6. **Nested Dictionaries**:
            If you need finer control over dictionary formats, you can replace `dict` with `{<string1_constant>: <type spec1>, <string2_constant>: <type spec2>, ...}`. The nested key type `<type spec1>` and value type `<type spec2>` can be any type conforming to the rules in this document (provided it is supported by Python; for example, `{[1]: 1}` will error in Python).

            3.6.1 **Fixed Length Dictionary**:
            e.g., `{"math": int, "english": int}` indicates a dictionary with two key-value pairs, where the keys must be the constants "math" and "english", and the corresponding values are basic types `int` and `int`. **Note: The dictionary length is fixed! Furthermore, key names must be constant strings; do not use the `{str: int}` syntax! If you need a variable-length dictionary, use the `Homogeneous Key, Homogeneous Value Dictionary Type`.**

            3.7. **Annotations (Description)**:
            You can insert annotation descriptions into the types defined in this document. The method is to replace the type `<type spec>` with `(<description>, <type spec>)`. For example: `("age", int)` is equivalent to `int`; `{"math": ("score", int)}` is equivalent to `{"math": int}`. Type annotations support nesting; you can insert annotations wherever appropriate. These annotations will be provided to the large model, enabling it to reply more accurately. It is recommended to use them frequently.

        For example, you can generate the following code for every step:
        [Example Start]
        # Step 1: Preparation
        dynasty_info_table = []

        # Step 2: Get dynasty list
        dynasty_list = llm.query(
            "Please list all major dynasties in Chinese history, such as Qin, Han, Tang, Song, Yuan, Ming, Qing, etc.",
            returns=[('dynasty list', list[str])]
        )

        # Step 3: Traverse each dynasty and emperor
        for current_dynasty in dynasty_list:
            
            # 3.1 Get all emperors of current dynasty
            emperor_list = llm.query(
                f"Please list all emperor names of {current_dynasty}",
                returns=[('emperor list', list[str])]
            )
            
            # 3.2 Traverse each emperor of current dynasty
            for current_emperor in emperor_list:
                # 3.2.1 Detailed query
                prompt = f"Please provide detailed information about '{current_emperor}' of '{current_dynasty}', specifically including: 1. Their commonly used era names; 2. Reign start and end dates; 3. Major historical events during their reign; 4. Corresponding Western historical periods or major countries during their reign; 5. Representative Sino-Western cultural exchange events of this period."
                reign_name, reign_period, main_events, contemporary_west, cultural_exchange = llm.query(
                    prompt,
                    returns=[
                        ("era name", str), 
                        ("reign period", str), 
                        ("major events", str), 
                        ("contemporary Western history", str), 
                        ("Sino-Western cultural exchange events", str)
                    ]
                )
                
                # 3.2.2 Add to data table
                dynasty_info_table.append({
                    "Dynasty": current_dynasty,
                    "Era Name": reign_name,
                    "Emperor": current_emperor,
                    "Reign Period": reign_period,
                    "Major Events": main_events,
                    "Contemporary Western History": contemporary_west,
                    "Sino-Western Cultural Exchange Events": cultural_exchange
                })
                
        [Example End]

        **Note**:
        1. `llm.query` is extremely useful. If you encounter irregular data items, you can directly call `llm.query` to have the large model organize them into regular data. Or, if there are calculation problems that are hard to solve, you can call `llm.query` to ask the large model to solve them.

**Note:**

1. When using the above framework, it is best to proceed step-by-step; do not click a large number of items in one go.
2. Successful code execution does not imply it meets expectations. Often `mobile.click/input` may fail. We suggest using `check` to set interface checkpoints to ensure the interface after each operation meets expectations.
3. You don't have to call a keyboard to input text, but use the `mobile.input(...)` method directly.
4. If you want to open an APP, you should use `mobile.start_app(app_name)` to open it, instead of trying to find the APP by swiping.
5. If `mobile.start_app(app_name)` fails to open the app, it may be because the task stack in the current device is not cleared. Try using multiple returns (using `mobile.back()`) to clear the current task stack, and then attempt to open the app again.
6. If you need to turn a system setting on or off, you need to first write a conditional check to see if it is in the expected state. If it is not in the expected state, then operate the switch. This ensures the setting is correctly opened or closed and avoids maloperation. For example:

```python
judge = mobile.check("the airplane mode is off.")
if judge:
    # Here you need to click the switch based on the specific view description
    mobile.click(view_description="Airplane mode switch")
```

8. The mobile phone is configured via a virtual mirror; its date and time are different from the real-world date and time!! The dates referred to in the Current Task are based on the **date on the mobile phone**, not the date in reality! Therefore, please make sure to emphasize using the date and time functions on the mobile phone to check today's date and time, rather than getting the real date via Python's `datetime` module! After obtaining "today's" date from the mobile phone, you can then use the `datetime` module to calculate other dates based on the mobile phone's "today".

For example, the following method is **wrong**:

```python
import datetime
today_date = datetime.date.today() # WRONG, because this gets the time on the Python executor, not the time in the mobile virtual machine! You should stick to the time on the mobile phone!
```

The following method is **correct**:

```python
today = ... # Get today's date from the mobile phone.
# Only then can you use the datetime library to calculate dates.
```

9. For any task involving viewing, deleting, or modifying entries, you **must click into the entry to view detailed information**, rather than simply looking at it in the thumbnail/list view and finishing. Thumbnails usually contain ellipses or "More..." prompts, reminding you that you need to click in to see the content because the information is truncated and you cannot see it completely from the outside. **Remember this! Failing to do so will result in failure!**
10. Regarding dates, "this {day_of_week}" refers to the {day_of_week} that has not yet arrived. For example, if today is Saturday, "this Tuesday" means the *next* upcoming Tuesday, not the Tuesday that has already passed. Therefore, be very clear when calculating dates to avoid errors.
11. If you find that the same `mobile.click/input` operation is consistently invalid after multiple attempts, you can try changing its `view_description` parameter. Make the `view_description` more detailed and specific to make its scope more precise, avoiding ambiguity or errors in the positioning process.

Here are some examples for you:

[Example 1 Start]
```python
ai_paper_table = []
mobile.start_app(app_name="Baidu")
# Use mobile.input directly and provide view_description
mobile.input(view_description="Search box", text="AI")
assert mobile.check(description="AI has been input into the search box.")  # check_result should be True
# Use mobile.click directly and provide view_description
mobile.click(view_description="Baidu Search Button")
assert mobile.check(description="Search result page for `AI`")  # check_result should be True
```
[Example 1 End]

[Example 2 Start]
```python
mobile.start_app(app_name=some_app)
...
# Swipe the post list
mobile.swipe_upward(view_description="<Title of a post at the bottom>", distance=400) # First determine the swipe start point as a post at the bottom, then swipe upward to view posts further down. Note: You must write the specific title name.
# Swipe the post list until the target post is found, max attempts 10. This is smarter and can swipe continuously.
mobile.swipe_until(view_description="<Title of a post at the bottom>", expected_desc="Hello world", towards='up', max_retry=10) # Swipe downward (content moves up) until the post named 'Hello world' is found. Note: You must write the specific title name as the start point.
post_list = llm.query(mobile.take_screenshot(), "List the posts currently displayed on the screen", returns=("Post list", list[str])) # Using llm.query combined with mobile.take_screenshot() allows for Q&A about the screen interface.
```
[Example 2 End]

[Example 3 Start]
```python
mobile.start_app(app_name=some_app)
...
# Drag the bar to the far right
mobile.swipe_rightward(view_description="the right part of the bar", distance=400) # Position the swipe center on the right side and drag it a long distance to the right to ensure it slides to the far right.
```
[Example 3 End]

**Remember again:**

1. If you want to open an APP, you should use `mobile.start_app(app_name)` to open it, instead of trying to find the APP by swiping.
2. The mobile phone is configured via a virtual mirror; its date and time are different from the real-world date and time!! The dates referred to in the Current Task are based on the **date on the mobile phone**, not the date in reality! Therefore, please make sure to emphasize using the date and time functions on the mobile phone to check today's date and time, rather than getting the real date via Python's `datetime` module! After obtaining "today's" date from the mobile phone, you should use the `datetime` module to calculate other dates based on the mobile phone's "today".

For example, the following method is **wrong**:

```python
import datetime
today_date = datetime.date.today() # WRONG, because this gets the time on the Python executor, not the time in the mobile virtual machine! You should stick to the time on the mobile phone!
```

The following method is **correct**:

```python
today = ... # Get today's date from the mobile phone.
# Only then can you use the datetime library to calculate dates.
```

3. For any task involving viewing, deleting, or modifying entries, you **must click into the entry to view detailed information**, rather than simply looking at it in the thumbnail/list view and finishing. Thumbnails usually contain ellipses or "More..." prompts, reminding you that you need to click in to see the content because the information is truncated and you cannot see it completely from the outside. **Remember this! Failing to do so will result in failure!**
4. If you find that the same `mobile.click/input` operation is consistently invalid after multiple attempts, you can try changing its `view_description` parameter. Make the description more detailed and specific to make its scope more precise, avoiding ambiguity or errors in the clicking process.

{current_date_info}
\end{lstlisting}

\end{document}